\documentclass{article} 
\ifdefined\XeTeXrevision\else\pdfoutput=1\fi
\usepackage[final]{colm2026_conference}

\usepackage{microtype}
\usepackage{hyperref}
\usepackage{url}
\usepackage{xurl}
\usepackage{needspace}
\makeatletter
\renewcommand\@makefnmark{\smash{\hbox{\textsuperscript{\normalfont\@thefnmark}}}}
\makeatother
\usepackage{booktabs}

\newcommand{\modelname}{\textsc{CAP}}
\usepackage{latexsym}
\usepackage[T1]{fontenc}
\usepackage[utf8]{inputenc}
\usepackage{inconsolata}
\usepackage{graphicx}
\usepackage{booktabs}
\usepackage{amsmath}
\usepackage{enumitem}
\usepackage{algorithm}
\usepackage{algorithmic}
\usepackage{xcolor}
\usepackage{multirow}
\usepackage{wrapfig}
\usepackage{siunitx}
\usepackage{caption}
\usepackage{pifont}
\usepackage{marvosym}
\usepackage{longtable}
\usepackage{booktabs}
\usepackage{tcolorbox}
\tcbuselibrary{breakable}
\tcbset{fonttitle=\bfseries, colframe=gray!55!black, colback=gray!6!white}
\usepackage{colortbl}
\usepackage{placeins}

\definecolor{errGray}{gray}{0.4}
\newcommand{\res}[2]{#1{\,\textcolor{errGray}{\scriptsize$\pm$#2}}}

\definecolor{myorange}{HTML}{E88A00}
\definecolor{myblue}{HTML}{1A56C4}
\definecolor{mypurple}{HTML}{9900ff}

\definecolor{myyellow}{HTML}{fff9eb}
\definecolor{myred}{HTML}{C0392B}
\definecolor{mygreen}{HTML}{34692E}
\definecolor{paperblue}{HTML}{077dea}
\definecolor{pergreen}{HTML}{eaf3ea}

\definecolor{checkgreen}{RGB}{34,139,34}
\definecolor{crossred}{RGB}{178,34,34}
\newcommand{\cmark}{\textcolor{checkgreen}{\ding{51}}}
\newcommand{\xmark}{\textcolor{crossred}{\ding{55}}}

\definecolor{darkblue}{rgb}{0, 0, 0.5}
\hypersetup{colorlinks=true, citecolor=darkblue, linkcolor=darkblue, urlcolor=darkblue, hyperfootnotes=false}

\title{\textcolor{myred}{C}\textcolor{myorange}{A}\textcolor{myblue}{P}: A Scalable Benchmark for Evaluating \textcolor{myred}{C}ross-Site Browser Agents with Complex \textcolor{myorange}{A}ctions and \textcolor{myblue}{P}erception}

\author{
Zejun Xu$^{*\dagger\,1,4}$ \quad
Taiyi Chen$^{*\,2,5}$ \quad
Jin Li$^{*\,5}$ \quad
Yongtong Gu$^{*\,3}$ \\
Qi Cheng$^{\dagger\,4}$ \quad
Aixuan Lv$^{4}$ \quad
Shuai Zhu \quad
Pengfei Zhu$^{1}$ \\
Kaichen Yang$^{\dagger\,4}$ \quad
Boyu Sun$^{3}$ \quad
Yixian Yang$^{1}$ \quad
Mulong Xie$^{4}$ \\
Xin Liu$^{1}$ \quad
Dagang Li$^{\text{\Letter}\,1}$ \quad
Xiaoteng Ma$^{\text{\Letter}\,2}$ \quad
Hongru Wang$^{\text{\Letter}\,6}$ \\
\small
\textsuperscript{1}Macau University of Science and Technology \quad
\textsuperscript{2}Tsinghua University \\
\textsuperscript{3}Southeast University \quad
\textsuperscript{4}FellouAI \quad
\textsuperscript{5}ARGUS Lab \quad
\textsuperscript{6}The University of Edinburgh \\
\texttt{zejunx@student.must.edu.mo} \\
\url{https://warriorxu0302.github.io/CAP-Bench/}
}

\begin{document}

\maketitle

\begingroup
\renewcommand{\thefootnote}{}
\footnotetext{$^{*}$Equal contribution.}
\footnotetext{$^{\dagger}$Work done during internship at FellouAI.}
\footnotetext{$^{\text{\Letter}}$Corresponding authors: Dagang Li (\texttt{dgli@must.edu.mo}), Xiaoteng Ma (\texttt{pony.xtma@gmail.com}), and Hongru Wang (\texttt{hongru.carrywang@gmail.com}).}
\endgroup

\begin{abstract}
Large language models are increasingly deployed as autonomous agents that interact with the web through browsers. While recent progress has been driven by benchmarks that evaluate end-to-end task success, these evaluations largely overlook two fundamental sources of difficulty in real web browsing: \textit{complex actions over rich user interfaces} and \textit{visual perception of dynamically rendered content}, especially in workflows that span multiple websites. We introduce \modelname, a scalable benchmark for evaluating browser agents on cross-site, human-like web tasks that require non-trivial UI interactions and visual understanding. Specifically, we adopt a decomposition-and-recomposition pipeline that first abstracts each website into a structured \emph{site card} capturing user-facing functions, complex execution operations, and perceptual requirements, and then recomposes these components into realistic cross-site workflows. Each task is therefore grounded in multiple specific operations on each website, enabling fine-grained diagnosis. Built on this framework, we construct 420 tasks across 108 real-world websites and 24 domains under careful quality control. Experiments on state-of-the-art browser agents using our verifiable agent-as-a-judge evaluation framework show low success rates and reveal that perception-heavy interactions remain a major bottleneck, exposing substantial gaps between current agents and real-world web browsing demands.
\end{abstract}

\section{Introduction}
\label{sec:intro}

Large language models increasingly act as autonomous agents that perceive, reason, and execute actions in real-world environments~\citep{wang2024survey,xi2023rise,acikgoz2025desideratum,wang2025position}. Among the interfaces these agents operate on, the web browser is especially important: it is the universal gateway to online information, services, and applications~\citep{yoran2024assistantbench,gou2025mind2web}. Browser agents such as OpenAI's Operator, Google's Project Mariner, Anthropic's computer use, and commercial products like Manus have appeared in quick succession~\citep{hu2025osagents}.

Yet real-world browser tasks are far more demanding than current benchmarks suggest. Consider a seemingly simple request from our benchmark: \emph{choosing a laptop}. A user might filter by specs and budget on BestBuy, compare stock availability across Amazon and Target, interpret star ratings and parse negative reviews for recurring complaints, and watch video reviews on YouTube, all within a single workflow. Most existing benchmarks fall short of capturing such realistic demands. They either focus on single-site interactions with simple operations such as clicking links or entering text~\citep{deng2023mind2web,zhou2023webarena}, or address cross-site tasks but typically evaluate agents only on final answers without systematically measuring the difficulty of the underlying interactions~\citep{yoran2024assistantbench,mialon2023gaia}. As a result, they offer limited insight into \textit{how} and \textit{why} agents fail in practice. In contrast, modern web applications present two intertwined challenges that existing evaluations largely overlook: 1) \emph{execution complexity}: completing a task often requires non-trivial interactions (e.g., dragging map regions, manipulating date-range sliders, or navigating nested dropdown menus) that go well beyond simple clicking and typing; and 2) \emph{perception complexity}: users must interpret rich visual content including charts, tables, images, and dynamically rendered elements to extract actionable information. Compounding these challenges, realistic workflows demand such complex interactions across multiple websites.

\begin{figure}[t]
\begin{center}
\includegraphics[width=0.78\linewidth]{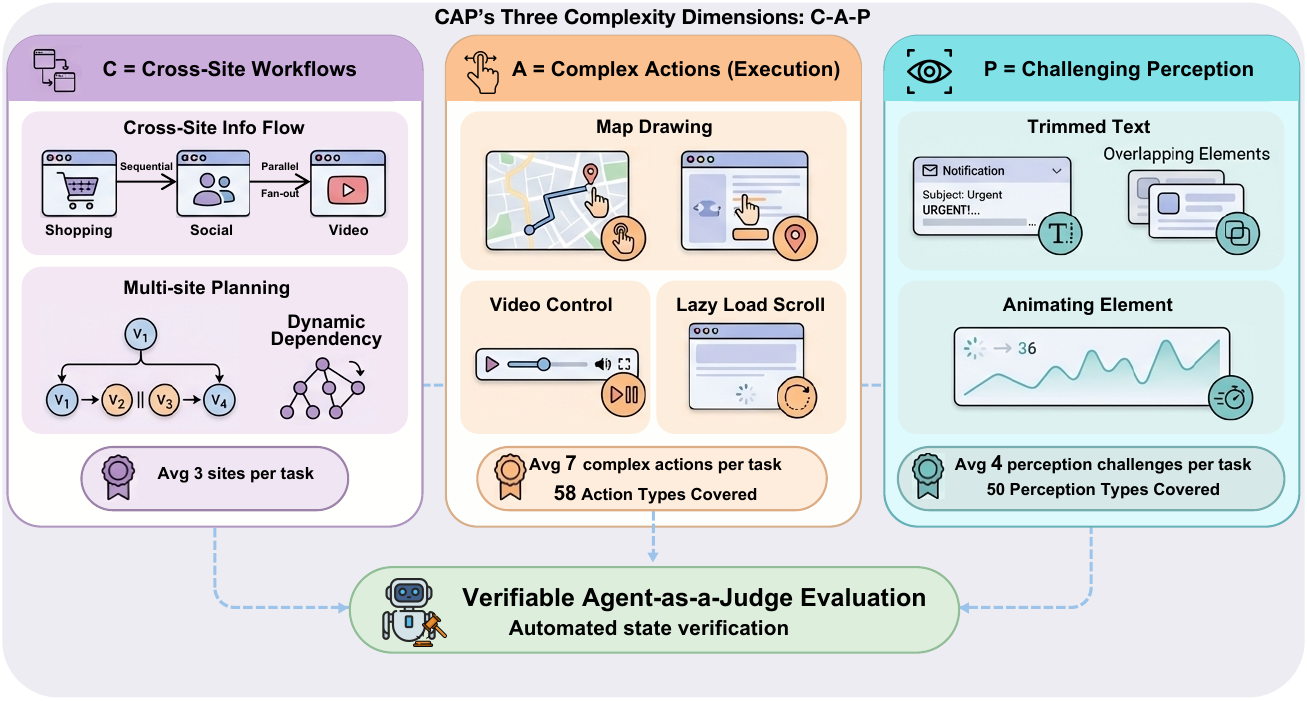}
\end{center}
\caption{Overview of the CAP framework. CAP targets three sources of difficulty in real-world web browsing (\textbf{C}ross-site workflows, complex \textbf{A}ctions, and challenging \textbf{P}erception) and evaluates agents through a verifiable agent-as-a-judge framework with explicit execution and perception checkpoints.}
\label{fig:intro_pipe}
\end{figure}

We introduce \textbf{CAP}, a scalable benchmark designed to evaluate browser agents under realistic web browsing conditions. As illustrated in Figure~\ref{fig:intro_pipe}, CAP targets three sources of difficulty central to human web use: \textbf{C}ross-site workflows, complex \textbf{A}ctions, and challenging visual \textbf{P}erception. To enable fine-grained grounding and diagnosis, we adopt a \emph{decomposition-and-recomposition} strategy. We first decompose each website into a structured \emph{site card} that catalogs what the site can do, what complex actions it requires (e.g., map dragging, video control), and what perceptual challenges it presents (e.g., interpreting charts, handling overlapping elements). We then recompose these building blocks into coherent cross-site task proposals through affinity-guided sampling, and finally instantiate them into concrete tasks with automatically generated evaluation criteria.

This design supports both expansion and diagnosis. New websites can be added by annotating their site cards, after which they are automatically integrated into tasks alongside existing sites without manual task engineering. At the same time, because each task is composed of explicitly annotated execution and perception components, evaluation can be grounded in explicit checkpoints rather than end results alone. Building on this, we propose a \emph{verifiable} agent-as-a-judge framework, where each task is automatically converted into a hierarchical rubric tree whose leaf nodes correspond to independently checkable execution and perception criteria, distinct from prior agent-as-a-judge approaches that rely on holistic LLM judgments over final outputs~\citep{gou2025mind2web}.

Our contributions are as follows:

\begin{itemize}
    \item We present CAP, a benchmark of 420 cross-site tasks spanning 108 real-world websites across 24 functional domains, with an average of 7 complex execution operations and 4 perception challenges per task.
    \item We develop a fully automated \emph{task synthesis} pipeline that decomposes websites into structured site cards and recomposes them into diverse, difficult cross-site tasks, enabling continuous benchmark expansion.
    \item We propose a verifiable agent-as-a-judge \emph{evaluation} framework that automatically generates hierarchical rubrics from task structure, providing fine-grained diagnosis of execution and perception capabilities.
    \item We evaluate eight state-of-the-art browser agent systems: even the best achieves only 8.0\% success rate, with perception-heavy interactions emerging as the major bottleneck.
\end{itemize}
\section{Related Work}
\label{sec:related_work}
\begin{table}[b]
    \centering
    \resizebox{\textwidth}{!}{%
    \begin{tabular}{@{}lccccccc@{}}
    \toprule
    \textbf{Benchmark} & \textbf{\# Tasks} & \textbf{Horizon} & \textbf{Dynamic} & \textbf{Cross-site} & \textbf{Exec. C.} & \textbf{Perc. C.} & \textbf{Evaluation} \\
    \midrule
    Online-Mind2Web~\citep{xue2025illusion}      & 300 & Short & \cmark  & \xmark & \xmark & \xmark  & LLM-as-a-Judge\\
    WebVoyager~\citep{he2024webvoyager}          & 643 & Short & \cmark  & \xmark & \xmark & \xmark  & LLM-as-a-Judge\\
    GAIA~\citep{mialon2023gaia}                  & 466 & Med. & \xmark & \xmark & \xmark & \xmark  & Answer Match\\
    AssistantBench~\citep{yoran2024assistantbench} & 214 & Med.  & \xmark & \xmark & \xmark & \xmark  & Answer Match\\
    Mind2Web2~\citep{gou2025mind2web}           & 130 & Long  & \cmark  & \cmark  & \xmark & \xmark  & Agent-as-a-Judge\\
    \midrule
    \rowcolor{gray!12}
    \textbf{\modelname\ (Ours)} & 420 & Long  & \cmark  & \cmark  & \cmark  & \cmark   & Verifiable Agent-as-a-Judge \\
    \bottomrule
    \end{tabular}%
    }
    \caption{Comparison with existing web browsing benchmarks. Horizon indicates task length, Dynamic indicates whether answers are time-varying, Cross-site indicates cross-website navigation, and Exec./Perc. Complex indicate requirements for complex execution and perception operations.}
    \label{tab:comp}
\end{table}

\paragraph{Benchmarks.}
Browser-based agents have been studied as a general interface for real web tasks, motivating benchmarks that address increasingly diverse functional demands~\citep{deng2023mind2web, zhou2023webarena, wang2024appbench, webbench2025}. Early benchmarks focused on programmatic interaction and structural information extraction within DOM-based environments~\citep{liu2018reinforcement, chen2021websrc}; later work introduced richer multimodal signals such as vision, enabling agents to operate on rendered pages while raising new evaluation challenges~\citep{xu2025turkingbench, song2025bearcubs}. Benchmarks have likewise moved from sandboxed simulators with fixed snapshots toward real-world, open environments that test agent robustness against the stochastic nature of the live web~\citep{pan2024webcanvas, kara2025warex, lu2025transbench}, and from short interactions toward long-horizon, multi-step tasks that demand sustained reasoning and long-term memory over extended trajectories~\citep{gou2025mind2web}. Yet as Table~\ref{tab:comp} shows, no existing benchmark jointly covers three dimensions of realistic browser use: cross-site workflows, complex execution operations, and challenging perceptual demands. Even benchmarks that support cross-site tasks~\citep{gou2025mind2web} score agents only on final answers rather than diagnosing where execution or perception breaks down. In contrast, CAP annotates and evaluates both execution and perception complexity within long-horizon, cross-site workflows.

\paragraph{Evaluation Methodologies.}
Evaluation methods for web agents trade off \emph{precision} against \emph{scalability}. \textbf{Rule-based graders}~\citep{mialon2023gaia, yoran2024assistantbench} use deterministic comparators over canonicalized outputs. They offer high precision and reproducibility, but require task-specific verification scripts whose authoring cost grows linearly with benchmark size, limiting scalability. \textbf{LLM-as-a-judge} approaches~\citep{he2024webvoyager, xue2025illusion} address this scalability bottleneck by applying rubric-based scoring with a general-purpose model. However, observing only the static final output, they cannot reliably verify multi-step claims against the underlying web state, leading to known false-positive issues on long-horizon tasks. \textbf{Agent-as-a-judge} methods~\citep{gou2025mind2web} instead shift from passive assessment to active verification (e.g., fetching and examining cited sources), at the cost of higher computational overhead. Yet existing agent-as-a-judge frameworks still judge only the final output and do not localize failures to specific execution or perception steps. Our \emph{verifiable} agent-as-a-judge framework extends this line of work by decomposing each task into a hierarchical rubric tree of independently checkable execution and perception criteria, yielding per-criterion capability diagnosis rather than one end-to-end score.

\section{\modelname\ Construction Pipeline}
\label{sec:method}

\begin{figure}[t]
\begin{center}
\includegraphics[width=0.8\linewidth]{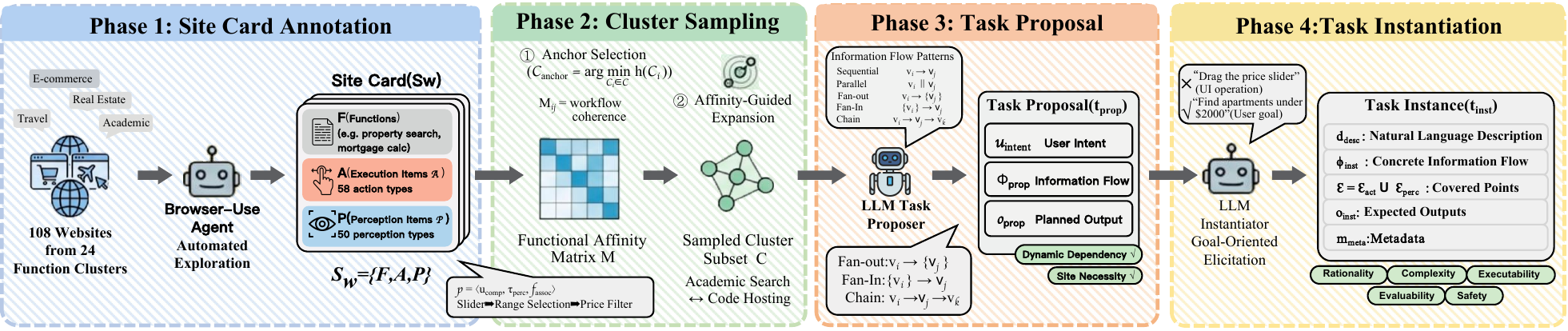}
\end{center}
\caption{Overview of the \modelname\ construction pipeline: (1) site card annotation decomposes websites into structured functional representations; (2) cluster sampling selects coherent website subsets by functional affinity; (3) task proposal generates abstract cross-site workflows; and (4) task instantiation produces concrete, verifiable task instances.}
\label{fig:task_construction_pipe}
\end{figure}

We introduce \modelname\ to evaluate agentic browsers on realistic cross-site tasks with non-trivial interactions. To construct the benchmark efficiently, we develop an automatic pipeline that generates such tasks.
As illustrated in Figure~\ref{fig:task_construction_pipe}, the pipeline consists of four phases: 1) site card annotation (\S~\ref{subsec:sitecard}); 2) cluster sampling (\S~\ref{subsec:cluster_sampling}); 3) task proposal (\S~\ref{subsec:task_proposal}); and 4) task instantiation (\S~\ref{subsec:task_inst}).

\subsection{Site Card Annotation}
\label{subsec:sitecard}

We use websites as the units for task construction. To ensure broad coverage of real-world web applications, we define 24 functional clusters $\mathcal{C} = \{C_1, C_2, \ldots, C_{24}\}$ commonly encountered in browser agent tasks, including e-commerce, travel booking, academic search, real estate, code hosting, video streaming, news media, and healthcare. Within each cluster $C_i$ we select high-traffic websites based on SimilarWeb rankings, yielding 139 candidates. We then exclude websites that block automated access or require mandatory login, since automated agents cannot reliably interact with them under standard configurations and including them would conflate agent capability failures with environment-level access barriers; 108 websites remain.

To capture what each website contributes to a task, we annotate it with a \emph{site card}, a structured summary that records what functions the website offers and what complex interactions those functions involve.
For each website $w$, the site card is defined as $s_w = \langle \mathcal{F}, \mathcal{A}, \mathcal{P} \rangle$. $\mathcal{F}$ denotes the set of user-facing functions. Each function $f \in \mathcal{F}$ describes what users can accomplish on the website (e.g., property search, mortgage calculation). $\mathcal{A}$ denotes the set of execution items, defined as non-trivial interactions such as map dragging and map zooming. Each execution item $a \in \mathcal{A}$ is defined as $a = \langle u_{comp}, \tau_{act}, f_{assoc} \rangle$, where $u_{comp}$ is the UI component, $\tau_{act}$ is the action type, and $f_{assoc} \in \mathcal{F}$ is the associated function.
$\mathcal{P}$ denotes the set of perception items, i.e., elements requiring visual understanding such as chart trend recognition and image content understanding. Each perception item $p \in \mathcal{P}$ is defined as $p = \langle u_{comp}, \tau_{perc}, f_{assoc} \rangle$, where $\tau_{perc}$ is the perception type. By linking execution items $a$ and perception items $p$ to functions $f_{assoc}$, we can phrase tasks in terms of user goals (e.g., ``find apartments under \$2000'') instead of UI operations (e.g., ``drag the price slider''), making the tasks better reflect real user scenarios.

To determine the scope of action types $\tau_{act}$ and perception types $\tau_{perc}$ that \modelname\ should cover, we ground our taxonomy in UI components, since modern web applications are built from them and each component constrains the interactions it affords. We adopt the Ant Design component library\footnote{\url{https://ant.design/}}, one of the most widely adopted enterprise-grade design systems (roughly 99k GitHub stars and over 3.5 million weekly npm downloads as of August 2026), as our reference framework.
Building on its taxonomy, we retain its core categories (\textit{Navigation}, \textit{Data Entry}, \textit{Data Display}, and \textit{Feedback}), exclude \textit{General} and \textit{Layout} components that primarily involve basic styling, and augment the taxonomy with \textit{Data Visualization}, \textit{Media}, and \textit{Editor} components that frequently appear in real-world websites but are not explicitly covered by the library. Based on this grounded component taxonomy, three domain experts in web development iteratively identify 64 action types and 56 perception types. The complete taxonomy is provided in Appendix~\ref{appendix:interaction_taxonomy}.

To annotate site cards $s_w$ efficiently, we use Browser-Use~\citep{browseruse2024}\footnote{\label{browser-use}\url{https://github.com/browser-use/browser-use}}, an open-source browser agent (prompt details in Appendix~\ref{appendix:prompt_sitecard}).
Given a website $w$ as input, the agent automatically explores the website, identifies user-facing functions $\mathcal{F}$, and catalogs execution items $\mathcal{A}$ and perception items $\mathcal{P}$, producing $s_w$ as output. The resulting 108 site cards cover 1,167 functions, with 58 of the 64 action types (90.6\%) and 50 of the 56 perception types (89.3\%) instantiated by at least one website. A manual error analysis of the automatically constructed site cards is reported in Appendix~\ref{appendix:quality_control}.

\subsection{Cluster Sampling}
\label{subsec:cluster_sampling}
In this stage, we select which $s_w$ to combine into cross-site tasks. Not all combinations make sense: browsing Zillow and then searching GitHub feels disjointed, whereas reading arXiv and then locating code on GitHub is a natural workflow. Since websites within a cluster share similar functionality, coherence between clusters generalizes to their member websites. We thus assess coherence once per cluster pair rather than per website pair; we term this cluster-level coherence \emph{functional affinity}.

To quantify functional affinity, we prompt a language model to assess pairwise cluster compatibility based on the plausibility of information flow scenarios, producing an affinity matrix $\mathbf{M}$ where $\mathbf{M}_{ij}$ represents the likelihood that clusters $C_i$ and $C_j$ form coherent workflows (prompt details in Appendix~\ref{appendix:prompt_affinity}). For instance, \emph{academic search} and \emph{code hosting} receive high $\mathbf{M}_{ij}$ since researchers frequently find papers and then locate implementations, whereas \emph{real estate} and \emph{code hosting} receive low $\mathbf{M}_{ij}$ due to the absence of natural information flow. Based on $\mathbf{M}$, we sample a cluster subset $\mathcal{C}^* \subseteq \mathcal{C}$ through two steps:

\paragraph{Step 1: Anchor selection.} We maintain historical usage statistics $h(C_i)$ that record how often cluster $C_i$ has appeared in generated tasks. We select the least-used cluster as the anchor, $C_{\text{anchor}} = \arg\min_{C_i \in \mathcal{C}} h(C_i)$, preventing over-representation of common domains.

\paragraph{Step 2: Affinity-guided expansion.} Starting from $\mathcal{C}^* = \{C_{\text{anchor}}\}$, we prompt a language model to select 3--6 clusters that work well with the anchor. We provide historical co-occurrence counts $h(C_i, C_j)$ for all cluster pairs and ask the model to favor high-affinity but rarely-combined clusters to diversify task coverage.

We pass the resulting subset $\mathcal{C}^*$ to the next stage for task proposal.

\subsection{Task Proposal}
\label{subsec:task_proposal}

The goal of this stage is to define what cross-site tasks should be performed over the websites that $\mathcal{C}^*$ covers. Manually authoring such tasks does not scale; the functions $\mathcal{F}$ recorded in site cards provide natural building blocks: by exposing what each website can do, they let a language model combine functions across sites into plausible workflows (prompts in Appendix~\ref{appendix:prompt_proposal}; flow patterns in Appendix~\ref{appendix:info_pattern}).

Given a sampled cluster subset $\mathcal{C}^*$ and the site cards $\{s_w\}$ of all websites $w$ belonging to clusters in $\mathcal{C}^*$, each generated proposal $t_{prop} = \langle u_{intent}, o_{prop}, \phi_{prop} \rangle$ contains three components: the user intent $u_{intent}$ describing who needs the task and why, the planned output $o_{prop}$ defining what the user expects to receive, and the information flow $\phi_{prop}$ specifying how data transfers between websites as a directed acyclic graph (DAG) over workflow steps.

Using this approach, we generate 600 task proposals (5 proposals per sampled cluster subset over 120 sampling rounds). Eight annotators manually review all proposals against two criteria: (i) the user intent reflects a meaningful real-world need rather than a contrived combination of features, and (ii) cross-site information dependencies are \emph{dynamically} obtained during execution rather than preset in the task description. Inter-annotator agreement on these criteria is high (Cohen's $\kappa = 0.81$), and 512 proposals (85.3\%) pass both checks and are forwarded to the instantiation stage.

\subsection{Task Instantiation}
\label{subsec:task_inst}
Given task proposals $t_{prop}$ from the previous stage, we instantiate each into a concrete benchmark task $t_{inst} = \langle d_{desc}, \phi_{inst}, \mathcal{E}, o_{inst}, m_{meta} \rangle$, comprising: a natural language description $d_{desc}$ phrased as a first-person user request, a concrete information flow $\phi_{inst}$ with specific entities and values, covered points $\mathcal{E} = \mathcal{E}_{act} \cup \mathcal{E}_{perc}$ denoting execution and perception items, expected outputs $o_{inst}$, and metadata $m_{meta}$ recording involved websites and functions. Each covered point $e \in \mathcal{E}$ is a tuple $e = (\textit{id}, \tau, d)$, where $\textit{id}$ is a unique identifier, $\tau \in \{\texttt{action}, \texttt{perception}\}$ indicates the type, and $d$ is a natural language description.

To generate $t_{inst}$, we select execution items $a \in \mathcal{A}$ and perception items $p \in \mathcal{P}$ from relevant site cards to include in $\mathcal{E}$, then prompt a language model to instantiate the abstract information flow $\phi_{prop}$ (\S\ref{subsec:task_proposal}) into $\phi_{inst}$ by binding abstract slots to concrete entities, and to concretize the planned output $o_{prop}$ into the expected outputs $o_{inst}$ (prompt details in Appendix~\ref{appendix:prompt_instantiation}). A key principle is \emph{goal-oriented elicitation}: descriptions express user goals (e.g., ``find what people are saying about this product'') rather than UI operations, so complex interactions arise from task requirements rather than explicit hints in the prompt. The outputs $o_{inst}$ serve both the user and verification: each covered point links to at least one output field reflecting its result, so the interaction can be checked indirectly.

After manual review for rationality, complexity, executability, evaluability, and topic safety (details in Appendix~\ref{appendix:quality_control}), we finalize 420 task instances (192 public, 228 private; we hold out the private split to mitigate test-set contamination in future evaluations). The benchmark spans 24 functional clusters and 108 websites, covering 405 functions and all 58 execution and 50 perception types instantiated in the site cards (90.6\% and 89.3\% of the full taxonomy). On average, each task involves 2 clusters and 3 websites. Detailed statistics are reported in Table~\ref{tab:detailed_stats} and Figure~\ref{fig:category_pie_charts} (Appendix~\ref{appendix:experimental_results}), and a complete task instance with its covered points and generated rubric is shown in Appendix~\ref{appendix:task_example}.

\section{Automated Evaluation Protocol}
\label{sec:evaluation}

Evaluating browser agents on composite, cross-site tasks faces two challenges: (1) final-answer accuracy alone cannot reveal whether agents struggle with complex UI operations (e.g., slider manipulation) or perceptual reasoning (e.g., trend identification); and (2) manually authoring evaluation scripts for hundreds of heterogeneous tasks is prohibitively expensive. To address both challenges, we propose a \emph{verifiable agent-as-a-judge} framework that uses the covered points annotated during task construction (\S\ref{sec:method}) to automatically generate fine-grained evaluation criteria. As illustrated in Figure~\ref{fig:eval_pipe}, the framework operates in three stages: (i) an LLM code generator synthesizes a task-specific Python evaluation script from the task description and covered points (\S\ref{sec:4_2}); (ii) the script materializes a hierarchical rubric tree whose leaves correspond to independently checkable execution and perception criteria (\S\ref{sec:rubric_tree}); and (iii) a judge agent traverses the tree, extracting and verifying claims from the agent's response to compute the scores (\S\ref{sec:score_comp}). Unlike prior agent-as-a-judge approaches that score final outputs as a whole~\citep{gou2025mind2web}, our framework grounds evaluation in explicit, type-annotated checkpoints.

\subsection{Problem Formulation}

Let $\mathcal{T} = \{t_1, t_2, \ldots, t_n\}$ denote the set of task instances in CAP. Recall from \S\ref{subsec:task_inst} that each task $t \in \mathcal{T}$ is associated with a set of covered points $E = E_{\text{act}} \cup E_{\text{perc}}$, each a tuple $e = (\textit{id}, \tau, d)$, where $E_{\text{act}}$ denotes \emph{execution points} (non-trivial interactive operations the task requires) and $E_{\text{perc}}$ denotes \emph{perception points} (elements requiring visual understanding). Given an agent's response $\mathcal{A}$ to task $t$, our goal is to compute a score $S(t, \mathcal{A}) \in [0, 1]$ that reflects both overall task completion and capability along the execution and perception dimensions.

\subsection{Evaluation Script Generation}
\label{sec:4_2}

\begin{figure}[t]
\begin{center}
\includegraphics[width=0.8\linewidth]{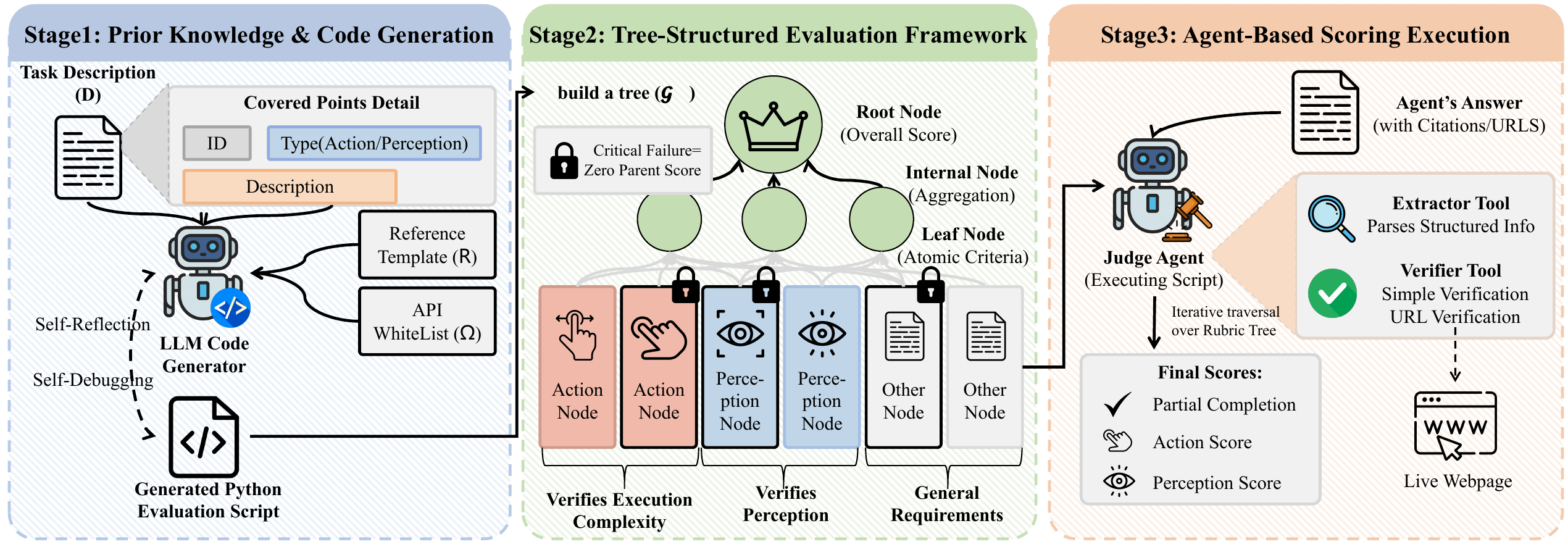}
\end{center}
\caption{Overview of the verifiable agent-as-a-judge evaluation framework: (1) an LLM synthesizes a task-specific evaluation script under an API whitelist and reference template; (2) the script materializes a hierarchical rubric tree with action, perception, and other leaf nodes; and (3) a judge agent performs extraction and URL-grounded verification to compute the four scores.}
\label{fig:eval_pipe}
\end{figure}

Because manual verification does not scale, we use an LLM-based code generation pipeline corresponding to Stage~1 of Figure~\ref{fig:eval_pipe}. Conditioned on the task description $D$ and the set of covered points $E$, the generator synthesizes a Python evaluation script $\mathcal{S}$ following a detailed instruction prompt (Appendix~\ref{appendix:eval_detail}). For safety and standardization, generation is constrained by two artifacts: an API whitelist $\Omega$ that restricts the script to a predefined set of safe verification primitives, and a reference template $\mathcal{R}$ of manually verified scripts that fixes the script's structural skeleton. The resulting $\mathcal{S}$ instantiates the rubric tree $G$ (defined in \S\ref{sec:rubric_tree}), encapsulating both leaf-node verification logic and hierarchical score aggregation. To improve robustness, execution failures trigger a self-debugging loop, followed by a self-reflection phase that validates logical correctness against a checklist of common edge cases (e.g., missing fields, malformed URLs, ambiguous thresholds).

\subsection{Rubric Tree Construction}\label{sec:rubric_tree}

Each evaluation script $\mathcal{S}$ implements a rubric tree $G = (V, \mathcal{E})$, a rooted tree where each node corresponds to a verification criterion. Leaf nodes $V_{\text{leaf}} \subset V$ represent atomic criteria, each producing a binary score $s(v) \in \{0, 1\}$; internal nodes aggregate scores from their children; and the root node $v_{\text{root}}$ yields the final task score. To enable capability diagnosis, we classify leaf nodes into three categories: \emph{action nodes} $V_{\text{act}}$ verify execution of complex UI operations, \emph{perception nodes} $V_{\text{perc}}$ verify extraction of visually-presented information, and \emph{other nodes} $V_{\text{other}}$ verify general requirements such as output format. Nodes are further marked as \emph{critical}, prerequisites whose failure invalidates downstream results and zeros out the parent score, or \emph{non-critical}, which permit partial credit. This classification is automatically determined by the LLM during script generation based on task-specific dependency analysis.

The rubric tree is constructed automatically from the task description $D$ and covered points $E$. Each covered point $e \in E$ maps to one or more leaf nodes according to its type: action-type points generate nodes in $V_{\text{act}}$, while perception-type points generate nodes in $V_{\text{perc}}$. This mapping enables indirect verification that agents complete the required execution and perception operations, without intrusive instrumentation of the agent's runtime. Because required operations are checked as critical leaves grounded in cited source pages, shortcut strategies such as answer guessing, plain search queries, or direct URLs fail verification (see Appendix~\ref{appendix:task_example}).

\subsection{Score Computation}
\label{sec:score_comp}

Given the rubric tree $G$ and an agent's response $\mathcal{A}$, scoring proceeds in two phases. Each leaf node $v \in V_{\text{leaf}}$ is first evaluated by a \emph{judge agent}, an LLM-based tool operating in two modes: \emph{extraction-and-verification}, which parses structured information from $\mathcal{A}$ and applies deterministic checks (e.g., numeric range, regex match, set membership), and \emph{URL verification}, which fetches cited web pages and validates that the supporting content matches the agent's claim. Internal nodes then aggregate child scores via a \emph{gate-then-average} strategy. Let $C(v)$ denote the children of an internal node $v$, partitioned into critical nodes $K(v)$ and non-critical nodes $N(v)$. The aggregated score is:
\begin{equation}
s(v) = 
\begin{cases}
0 & \text{if } \exists u \in K(v) : s(u) < 1 \\
\bar{s}_N & \text{if } \forall u \in K(v) : s(u) = 1 \text{ and } N(v) \neq \emptyset \\
1 & \text{otherwise}
\end{cases}
\end{equation}

where $\bar{s}_N = \frac{1}{|N(v)|} \sum_{u \in N(v)} s(u)$: any failed critical child propagates a zero up the tree (\emph{gating}), whereas non-critical children contribute their average once all critical prerequisites are satisfied (\emph{averaging}).

We report four metrics: (i) \textbf{Partial Completion}, the mean root-node score $s(v_{\text{root}})$ averaged over all tasks, capturing graded progress; (ii) \textbf{Success Rate}, the proportion of tasks achieving a \emph{perfect score}, defined as $s(v) = 1$ for every node $v \in V$, capturing end-to-end task success; (iii) \textbf{Action Score}, the mean leaf score over $V_{\text{act}}$, isolating execution capability; and (iv) \textbf{Perception Score}, the mean leaf score over $V_{\text{perc}}$, isolating perceptual capability (reported as \textbf{Complex-A} and \textbf{Complex-P} in \S\ref{sec:Experiments}). All metrics are normalized to $[0, 1]$. We further validate the judge against human judgments and across alternative judge backbones in Appendix~\ref{appendix:judge_validation}.
\section{Experiments}
\label{sec:Experiments}

\begin{table*}[htbp]
  \centering
  \renewcommand{\arraystretch}{1.1}
  \resizebox{\textwidth}{!}{%
  \begin{tabular}{l|cc|cc|cc}
    \toprule
    \multirow{2}{*}{\textbf{Method}} & \multicolumn{2}{c|}{\textbf{Core Performance (\%)}} & \multicolumn{2}{c|}{\textbf{Complexity (\%)}} & \multicolumn{2}{c}{\textbf{Output}} \\
    \cmidrule(lr){2-3} \cmidrule(lr){4-5} \cmidrule(lr){6-7}
     & \textbf{Partial Completion} & \textbf{Success Rate} & \textbf{Complex A} & \textbf{Complex P} & \textbf{Time (min)} & \textbf{Len. ($\times 10^3$)} \\
    \midrule
    
    \multicolumn{7}{l}{\textit{\textbf{Commercial Browser Agents}}} \\
    \midrule
    
    Genspark & \res{16.0}{2.0} & \res{5.0}{1.0} & \res{30.0}{2.1} & \res{28.0}{1.8} & 1.7 & 3.37 \\
    Manus    & \res{23.0}{3.0} & \res{8.0}{2.0} & \res{33.0}{2.5} & \res{29.0}{2.0} & 5.3 & 7.55 \\
    Dia      & \res{19.0}{2.0} & \res{4.0}{1.0} & \res{28.0}{1.5} & \res{23.0}{2.2} & \textbf{1.1} & 3.00 \\
    Fellou   & \res{24.0}{2.0} & \res{7.0}{2.0} & \res{28.0}{1.9} & \res{25.0}{2.1} & 22.1 & 3.47 \\
    Comet    & \textbf{\res{48.0}{3.5}} & \res{6.0}{1.5} & \textbf{\res{67.0}{4.2}} & \textbf{\res{58.0}{3.8}} & 6.5 & 104.36 \\

    \addlinespace[0.5em]
    \midrule
    \multicolumn{7}{l}{\textit{\textbf{Open Source Agents}}} \\
    \midrule
    
    GPT-5  & \res{15.0}{1.0} & \res{2.0}{1.0} & \res{29.0}{2.0} & \res{23.0}{1.5} & 15.3 & \textbf{2.95} \\
    Claude-4.5-Sonnet & \res{21.0}{3.0} & \res{5.0}{2.0} & \res{32.0}{2.4} & \res{32.0}{2.5} & 29.9 & 5.52 \\
    DeepSeek-V4-Flash & 25.0{\,\scriptsize\hphantom{$\pm$0.0}} & 2.9{\,\scriptsize\hphantom{$\pm$0.0}} & 33.7{\,\scriptsize\hphantom{$\pm$0.0}} & 34.2{\,\scriptsize\hphantom{$\pm$0.0}} & -- & -- \\

    \midrule

    \rowcolor{gray!12}
    Human & \res{35.0}{4.5} & \textbf{\res{10.0}{2.5}} & \res{35.0}{3.0} & \res{34.0}{2.8} & 16.3 & 71.42 \\

    \bottomrule
  \end{tabular}%
  }
  \caption{Main results on CAP benchmark. All performance metrics are reported in percentages (\%). Error bars (indicated in gray) represent standard deviation; dashes indicate values not recorded. Note that \textbf{GPT-5}, \textbf{Claude-4.5-Sonnet}, and \textbf{DeepSeek-V4-Flash} in the Open Source section refer to the \textit{Browser-Use} agent framework driven by the respective backbone models.}
  \label{tab:main_result_bw}
\end{table*}

\subsection{Setup}
\paragraph{Evaluated Systems.}
We evaluate both commercial and open-source browser agents on a fixed evaluation set of 198 CAP tasks\footnote{The 192-task public split of CAP is released in full, while the remaining 228 tasks are held out to mitigate test-set contamination; submissions on the private split are evaluated by the authors.} constructed in \S\ref{sec:method}.
We benchmark five commercial browser agent products, selected for their ability to handle complex multi-site interaction: Manus, Genspark, Fellou, Comet (Gemini~3.1~Pro), and Dia\footnote{\url{https://manus.im/}; \url{https://www.genspark.ai/}; \url{https://fellou.ai/}; \url{https://www.perplexity.ai/comet}; \url{https://www.diabrowser.com/}.}, all accessed through their official interfaces with default configurations during November--December~2025.
For open-source baselines, we evaluate Browser-Use\footref{browser-use} driven by three backbones (GPT-5, Claude-4.5-Sonnet, and the open-weights DeepSeek-V4-Flash) with default agent configurations and a maximum of 50 reasoning--action steps per task; sharing one framework, these rows isolate backbone capability.
Finally, we include a \textbf{Human baseline} produced by the twelve CS graduate students from our annotation pipeline (Appendix~\ref{appendix:quality_control}), with each task completed under a one-hour time limit and scored by the same judge and rubrics as the agents; it calibrates CAP's absolute difficulty under our strict conjunctive criterion rather than serving as a direct comparison target.

\paragraph{Evaluation Metrics.}
We report the four metrics defined in \S\ref{sec:score_comp}: \textbf{Partial Completion}, \textbf{Success Rate}, and the two capability scores, reported as \textbf{Complex-A} (action) and \textbf{Complex-P} (perception). For context on cost, we additionally report the average end-to-end \textbf{Time} per task (minutes) and the average output \textbf{Length} (thousands of tokens). All systems are scored by the same judge with a GPT-4o backbone; Appendix~\ref{appendix:judge_validation} validates the judge against human judgments and across four alternative judge backbones.

\subsection{Main Results}

\noindent\textbf{Finding 1: State-of-the-art browser agents perform poorly on complex execution and perception tasks.}
The main results are presented in Table~\ref{tab:main_result_bw}. Success rates are low across all evaluated systems: the best agent, Manus, reaches only 8.0\% Success Rate, and even the Human baseline achieves just 10.0\% full success and 35.0\% partial completion. We attribute the modest human ceiling to two factors: (i) tasks deliberately span 2--14 websites with average 7 complex actions and 4 perception challenges, beyond what most users reliably execute in an hour; and (ii) our perfect-score criterion requires \emph{every} execution, perception, and format leaf to be satisfied. A manual audit of failed human runs confirms this. Humans typically fail by omitting a single required constraint in a long workflow, not by being unable to perform individual steps. We therefore view Partial Completion as the more informative comparison metric.
Comet is an outlier: it leads on Partial Completion (48.0\%) and both complexity sub-scores, yet underperforms on Success Rate (6.0\%). This gap, together with its order-of-magnitude longer outputs (104k vs.\ roughly 3--8k tokens), suggests Comet gets the intermediate reasoning right but fails to consolidate it into a final answer satisfying all rubric leaves.

\noindent\textbf{Finding 2: Perception is a more severe bottleneck than execution for current browser agents.}
On Complex-A, Comet leads at 67.0\%, while every system drops on Complex-P (Comet 58.0\%, others 23.0--34.2\%). All five commercial agents and Browser-Use~+~GPT-5 score higher on Complex-A than on Complex-P (e.g., Comet 67.0\% vs.\ 58.0\%), whereas the Human baseline is essentially balanced (35.0\% vs.\ 34.0\%), as are the Claude-4.5-Sonnet and DeepSeek-V4-Flash backbones. Together with the near-zero \emph{value reading} score (\S\ref{subsec:complexity_analysis}), this asymmetry shows that visual reasoning over rendered content, rather than UI manipulation per se, is the dominant failure mode of current agents.

\noindent\textbf{Finding 3: Extended reasoning correlates with task progress, but no agent matches human deliberation.}
Comet generates over 100k output tokens to reach its leading Partial Completion, an order of magnitude more than other agents (roughly 3--8k), and the Human baseline spends 16.3 minutes per task on average, whereas fast-responding agents (Genspark 1.7~min, Dia 1.1~min) sit at the bottom of the Partial Completion ranking. This is consistent with multi-site workflows requiring sustained reasoning that short-horizon agents cannot shortcut; still, compute alone is insufficient: Browser-Use~+~GPT-5 spends 15.3~min yet achieves only 15.0\% Partial Completion, echoing the view that acting efficiently matters more than acting more~\citep{wang2025acting}.

\subsection{Complexity Analysis}
\label{subsec:complexity_analysis}

\noindent\textbf{Finding 4: Fine-grained visual reading and UI state inference are where agents fail most.}
Figure~\ref{fig:category_bar_charts}(a-b) (Appendix~\ref{appendix:experimental_results}) reveals the hardest components in CAP. Among execution operations, \emph{node manipulation} (0.087) and \emph{date range selection} (0.110) score lowest, both involving precise pointer control over composite widgets. Among perception operations, \emph{value reading} achieves a near-zero score (0.000) and \emph{expansion state awareness} (0.092) is also severely impaired, indicating that extracting concrete values from rendered charts/tables and inferring whether a UI region is expanded remain unsolved for current agents (the near-zero value-reading score reflects attempted-but-failed extractions, not missing attempts).

To quantify task-level complexity, we define two metrics over the evaluation set: \textbf{raw-sum complexity}, the total count of execution plus perception leaves per task, and \textbf{normalized complexity}, which divides each count by its benchmark-wide maximum and sums the two ratios, yielding a value in $[0, 2]$.

In Figure~\ref{fig:complexity_comparison} (Appendix~\ref{appendix:experimental_results}), each task contributes one point per panel, plotting its complexity against the number of websites it involves, colored by its score in $[0, 1]$. We find that \textbf{task difficulty (score) is not monotonically explained by complexity quantity}: high- and low-scoring tasks distribute across the entire complexity range under both metrics, so what makes a task hard is not the \emph{number} of complex actions or perceptions but their \emph{kind}: a single chart-value-reading step can derail an otherwise simple workflow.

\subsection{Cross-site Analysis}

\noindent\textbf{Finding 5: Domain-specific sites with specialized data structures are disproportionately hard.}
The per-website scores in Figure~\ref{fig:category_bar_charts}(c) (Appendix~\ref{appendix:experimental_results}) vary substantially: \texttt{cdc.gov} (0.033) and \texttt{google.com/flights} (0.048) are the hardest, while \texttt{accuweather.com} (0.139) and \texttt{wikipedia.org} (0.131) are relatively tractable. The hardest sites present information through specialized visualizations or interactive widgets and often reveal the relevant content only after non-trivial filtering, dragging, or chart hovering, whereas easier sites expose the target information as nearly static text.

Figure~\ref{fig:complexity_comparison} also shows a moderate positive correlation between website count and task complexity (normalized: $r{=}0.496$; raw-sum: $r{=}0.486$). This correlation is, however, dominated by tasks involving 2--4 websites ($n{=}41, 80, 22$); buckets with $\geq 5$ websites contain few samples ($n{\leq}10$) and are illustrative only.
\section{Conclusion}

We introduced \modelname, a scalable benchmark of 420 tasks over 108 real-world websites that evaluates browser agents on \textbf{C}ross-site workflows, complex \textbf{A}ctions, and challenging visual \textbf{P}erception. Tasks are built by a decomposition-and-recomposition pipeline over structured site cards and scored by a \emph{verifiable agent-as-a-judge} rubric framework for fine-grained diagnosis. Across eight state-of-the-art systems, the best agent reaches only 8.0\% Success Rate, and perception consistently lags execution; visual perception, rather than UI manipulation, is where progress is most needed. We release the public split, construction pipeline, and evaluation framework\footnote{\url{https://github.com/WarriorXu0302/CAP-Bench}}; limitations are discussed in Appendix~\ref{app:limitations}.

\section*{Acknowledgments}
This work was supported by the Fund for the Development of Science and Technology (FDCT) of Macau (Grant No.~0010/2024/AGJ).

\section*{Reproducibility Statement}
We release the 192-task public split of \modelname\ in full\footnote{Dataset: \url{https://huggingface.co/datasets/Warrior0302/CAP-Bench}. Construction pipeline and evaluation framework: \url{https://github.com/WarriorXu0302/CAP-Bench}.}, together with site cards, task descriptions, rubric trees, evaluation scripts, and per-task annotation timestamps (with archived URLs where available); the remaining 228 tasks are held out to mitigate test-set contamination, and private-split submissions are evaluated by the authors via the \modelname\ leaderboard\footnote{\url{https://warriorxu0302.github.io/CAP-Bench/leaderboard.html}}. Because \modelname\ runs against the live web, absolute scores will drift as websites evolve: we mitigate this by annotating and evaluating each task within roughly one week of each other, by avoiding volatile targets at construction time, and by grounding scoring in the judge's source-URL verification. We recommend reporting the evaluation date alongside scores. Further limitations are discussed in Appendix~\ref{app:limitations}, and validation of the automated judge in Appendix~\ref{appendix:judge_validation}.

\bibliography{colm2026_conference}
\bibliographystyle{colm2026_conference}

\appendix

\counterwithin{figure}{section}
\counterwithin{table}{section}

\renewcommand{\thefigure}{\thesection.\arabic{figure}}
\renewcommand{\thetable}{\thesection.\arabic{table}}

\section{Limitations}
\label{app:limitations}

Several scope decisions keep the benchmark focused and interpretable, but each brings limitations. We discuss the main ones below, along with our mitigations.

\paragraph{Coverage of evaluated systems.}
Our experimental study evaluates eight representative browser agent systems rather than exhaustively covering all existing systems or model variants. This choice reflects CAP's focus on \emph{interaction-level} capabilities (complex UI execution and visual perception) rather than model scaling effects. Including a small number of strong commercial systems alongside widely used open-source agents lets us expose capability gaps that persist across architectures, but the absolute numbers reported here should be read as a snapshot of late~2025 systems rather than a final ranking; we will refresh results as new agents emerge.

\paragraph{Coverage of tasks and domains.}
While CAP spans 24 functional domains and 108 websites, it does not aim to cover every possible web interaction or long-tail domain. We deliberately emphasize interaction patterns that (i) commonly occur in real user workflows and (ii) admit automatic verification through our rubric framework. Tasks requiring real-money transactions, account-bound private data, or strong real-world side effects (e.g., actually purchasing items, sending emails) are excluded for safety and reproducibility, and are left to future work.

\paragraph{Reliability of the automated construction pipeline.}
Site cards, task proposals, and task instances are produced by LLM-driven agents and audited by human annotators (\S\ref{subsec:task_inst}, Appendix~\ref{appendix:quality_control}). Despite multi-stage quality control, automatically extracted site cards may miss rare or deeply-nested functions, and synthesized tasks may carry minor phrasing artifacts that no human author would write. Our manual error analysis of site cards (Appendix~\ref{appendix:quality_control}) finds 92.3\% of entries accurate, with residual errors concentrated in over-predicted action items that cannot propagate into tasks thanks to the manual execution check. Human auditing scales to the current 420 tasks but does not scale linearly to tens of thousands of tasks; this remains an open problem for any automated benchmark of this kind.

\paragraph{Residual noise in the verifiable agent-as-a-judge framework.}
Our evaluation framework reduces the cost of authoring per-task scripts by relying on LLM code generation and an LLM-based judge agent. Although we constrain the generator with an API whitelist, a reference template, and an iterative self-debug/self-reflection loop, the resulting scripts may still contain edge-case bugs, and the judge agent inherits the residual unreliability of its underlying LLM (e.g., when verifying claims against fetched web pages whose content has shifted between agent execution and judge inspection). We mitigate this by auditing the judge against human judgments on a 50-task sample (96\% verdict agreement, judge--human $\kappa = 0.84$; Appendix~\ref{appendix:judge_validation}), by checking that generated rubrics faithfully capture their tasks (96\%), and by reporting standard deviations across repeated runs in Table~\ref{tab:main_result_bw}, but we do not claim noise-free verification.

\paragraph{Live-web reproducibility.}
CAP runs against the live web, which makes the benchmark realistic but also non-stationary: website layouts, content, and feature availability change over time, and absolute scores reported here will drift. Construction avoids volatile targets (relative time windows instead of fixed dates, range filters over exact values, and multiple acceptable answers rather than one fragile target), annotation and evaluation are separated by roughly one week, and the judge's source-URL verification grounds scoring in what pages actually show at evaluation time; see the Reproducibility Statement for the release plan that lets future drift be audited. Still, exact score reproducibility across long time intervals is not guaranteed.

\clearpage
\section{Benchmark Statistics and Experimental Results}
\label{appendix:experimental_results}

\begin{table}[h]
    \centering
    \renewcommand{\arraystretch}{1.1}
    \small
    \begin{tabular}{@{}lcccc@{}}
    \toprule
    \textbf{Category} & \textbf{Types} & \textbf{Avg} & \textbf{Min} & \textbf{Max} \\
    \midrule
    Clusters        & 24    & 2    & 1   & 5  \\
    Websites       & 108   & 3  & 2    & 14   \\
    Functions      & 405  & 5  & 2    & 15   \\
    \midrule
    Execution      & 58   & 7  & 2    & 20  \\
    Perception     & 50   & 4  & 2    & 15  \\
    \bottomrule
    \end{tabular}
    \caption{Statistics of our benchmark. Types refers to the total distinct types across all queries. Avg, Min, and Max refer to the number of operations/elements per task.}
    \label{tab:detailed_stats}
    \vspace{-6mm}
\end{table}

\begin{figure}[h]
  \centering
  \includegraphics[width=0.62\linewidth]{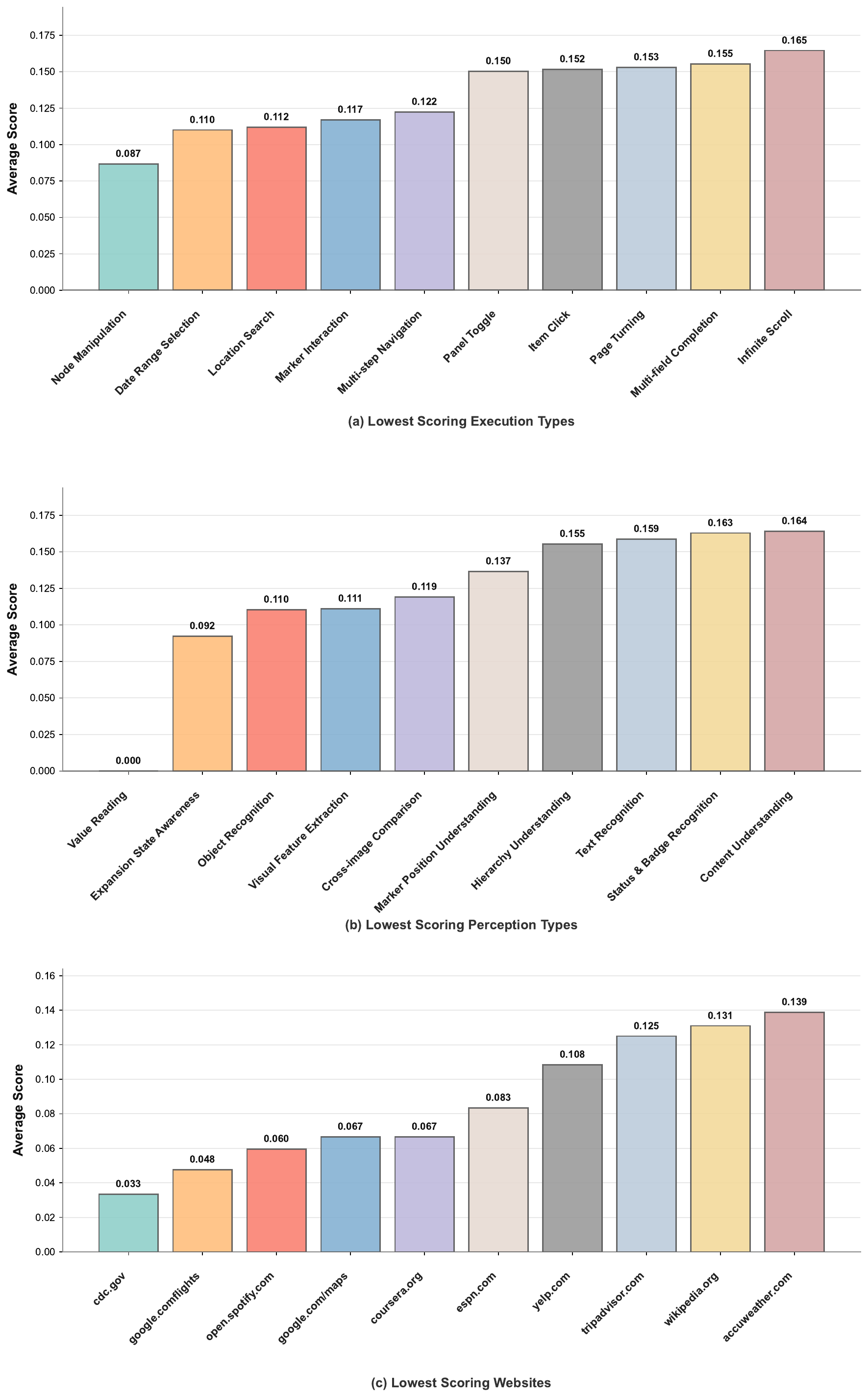}
  \vspace{-2mm}
  \caption{Distributions over the bottom-10 (by mean score) complex-execution types, complex-perception types, and websites.}
  \label{fig:category_bar_charts}
\end{figure}

\vspace{-3mm}

\begin{figure}[h]
  \centering
  \includegraphics[width=0.85\linewidth]{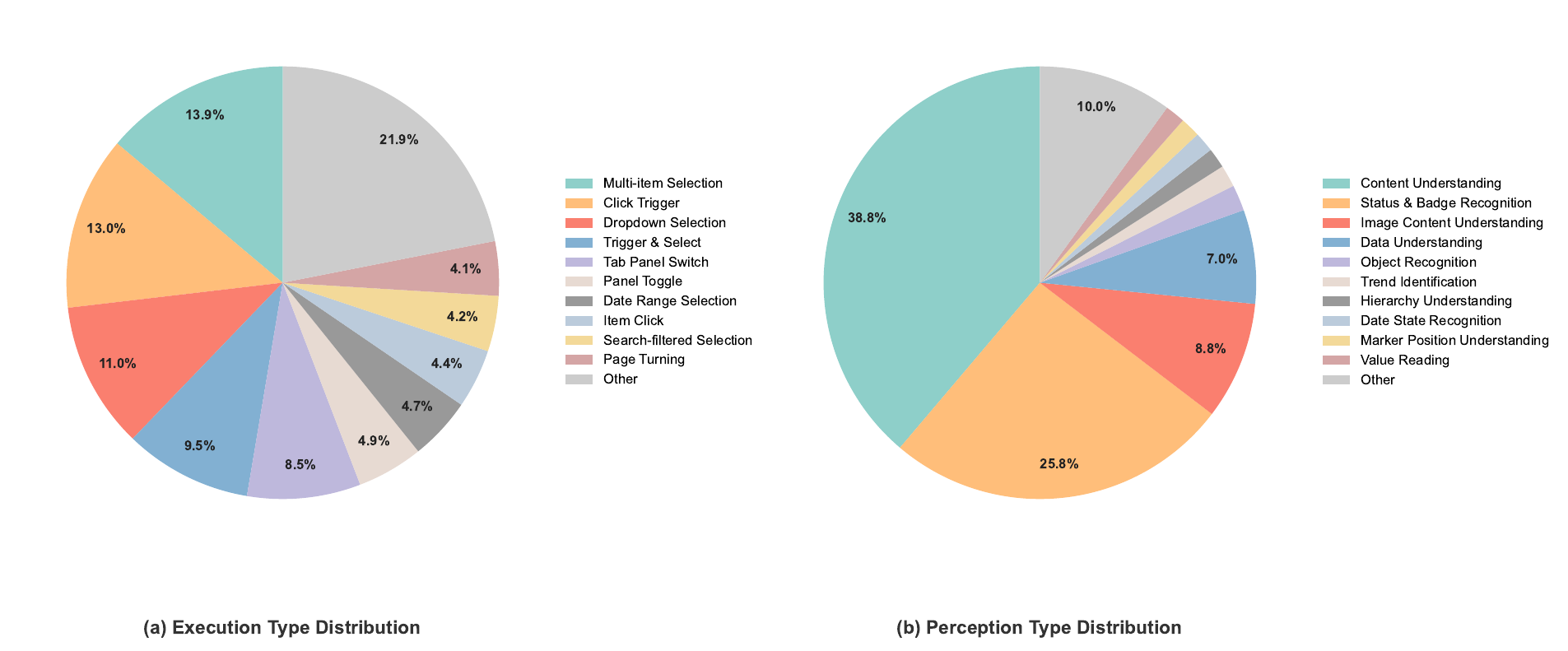}
  \vspace{-2mm}
  \caption{Distribution of (a) complex execution types and (b) complex perception types in our benchmark.}
  \label{fig:category_pie_charts}
\end{figure}

\begin{figure}[h]
  \centering
  \includegraphics[width=0.9\linewidth]{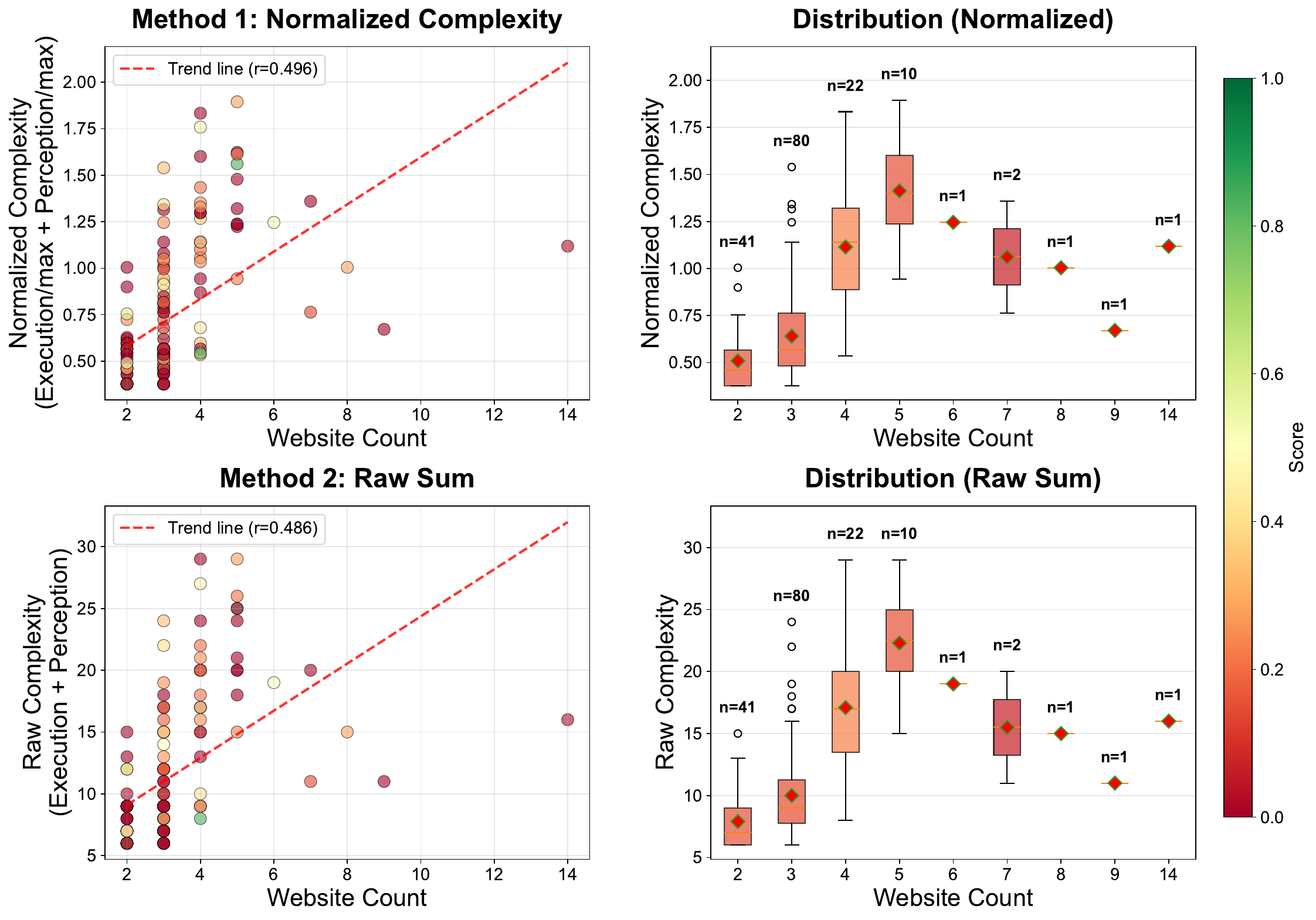}
  \caption{Task complexity as a function of the number of websites a task involves, under the two complexity definitions of \S\ref{subsec:complexity_analysis} (normalized and raw-sum). Left: each point is one task, with linear fits; point color encodes the task's evaluation score (red = low, green = high). Right: the same data as box plots grouped by website count, with the number of tasks $n$ annotated above each box.}
  \label{fig:complexity_comparison}
\end{figure}

\vspace{-3mm}

\FloatBarrier

\section{Taxonomy of Complex Interaction Types}
\label{appendix:interaction_taxonomy}

We define a taxonomy comprising 64 action types and 56 perception types. The taxonomy is grounded in the Ant Design component library, excluding basic categories (\textit{General}, \textit{Layout}) and extending with \textit{Data Visualization}, \textit{Media}, and \textit{Editor} to cover complex interactions in real-world applications. The complete lists are provided in Table~\ref{tab:action_types} and Table~\ref{tab:perception_types}.

\subsection{Action Types}
\label{appendix:action_types}

Action types ($\tau_{act}$) define the interactive operations that an agent must perform on UI components. We categorize these into seven primary categories: \textit{Navigation}, \textit{Data Entry}, \textit{Data Display}, \textit{Feedback}, \textit{Data Visualization}, \textit{Media}, and \textit{Editor}.

{
\small
\centering
\setlength{\tabcolsep}{5pt}
\begin{longtable}{@{}llp{2.9cm}p{6.1cm}@{}}
	\toprule
	\textbf{ID} & \textbf{Component} & \textbf{Action Type} & \textbf{Description} \\
	\midrule
	\endfirsthead
	
	\toprule
	\textbf{ID} & \textbf{Component} & \textbf{Action Type} & \textbf{Description} \\
	\midrule
	\endhead
	
	\midrule
	\endfoot
	
	\bottomrule
	\noalign{\vskip 6pt}
	\caption{Action types for UI components.}
	\label{tab:action_types} \\
	\endlastfoot
	
	\multicolumn{4}{l}{\textit{Navigation}} \\*
	AT01 & Dropdown & Trigger \& Select & Click or hover to trigger dropdown and select target item \\
	AT02 & Dropdown & Multi-level Navigation & Navigate through nested dropdown menus to locate target \\
	AT03 & Menu & Submenu Expansion & Expand multi-level submenus via hover or click (e.g., mega menu) \\
	AT04 & Pagination & Page Turning & Click page numbers or prev/next buttons to switch pages \\
	AT05 & Pagination & Page Jump & Input specific page number to jump directly \\
	AT06 & Steps & Multi-step Navigation & Execute prev/next transitions in wizard workflows \\
	AT07 & Tabs & Tab Panel Switch & Click tab labels to switch content panels \\
	AT08 & Tabs & Tab Bar Scrolling & Scroll tab bar to reveal hidden tab options \\
	AT09 & Breadcrumb & Path Navigation & Click breadcrumb items to navigate to target hierarchy \\
	\midrule
	
	\multicolumn{4}{l}{\textit{Data Entry}} \\*
	AT10 & Checkbox & Multi-item Selection & Select multiple checkboxes for combined filtering \\
	AT11 & Checkbox & Select All/Inverse & Execute select all, inverse selection, or deselect all \\
	AT12 & Cascader & Hierarchical Selection & Expand levels sequentially to complete selection (e.g., region) \\
	AT13 & Cascader & Search-based Selection & Search keywords to directly locate and select deep options \\
	AT14 & DatePicker & Single Date Selection & Open calendar panel and select date with quick options \\
	AT15 & DatePicker & Date Range Selection & Select start/end dates with shortcuts (e.g., last 7 days) \\
	AT16 & DatePicker & DateTime Selection & Select both date and specific time simultaneously \\
	AT17 & Form & Multi-field Completion & Identify and complete forms with multiple input types \\
	AT18 & Select & Dropdown Selection & Open dropdown list for single or multi-select scenarios \\
	AT19 & Select & Search-filtered Selection & Filter options via keywords then select target item \\
	AT20 & Slider & Single Value Drag & Drag slider to select a single value \\
	AT21 & Slider & Range Selection & Drag dual sliders to set value range (e.g., price range) \\
	AT22 & TimePicker & Time Selection & Select specific hour/minute via scroll or input \\
	AT23 & TreeSelect & Tree Node Selection & Expand tree structure and select target nodes \\
	AT24 & Upload & File Upload & Upload files via click or drag-and-drop \\
	\midrule
	
	\multicolumn{4}{l}{\textit{Data Display}} \\*
	AT25 & Calendar & Date Selection & Click to select target date in calendar view \\
	AT26 & Calendar & View Switch & Switch between different month or year views \\
	AT27 & Card & Click Trigger & Click card for details or action buttons (favorite, add to cart) \\
	AT28 & Card & Hover Trigger & Hover to reveal quick actions or additional information \\
	AT29 & Carousel & Slide Switch & Switch via arrows/indicators or drag gestures \\
	AT30 & Collapse & Panel Toggle & Click header to expand/collapse, including accordion mode \\
	AT31 & Image & Preview Operations & Open thumbnail for zoom, rotate, or image switching \\
	AT32 & Popover & Popover Interaction & Trigger popover via click/hover and perform actions within \\
	AT33 & Table & Column Sorting & Click header to sort by column in ascending/descending order \\
	AT34 & Table & Column Filtering & Use column filters to filter table data by conditions \\
	AT35 & Table & Row Operations & Select rows, expand details, or click inline action buttons \\
	AT36 & Tree & Node Manipulation & Expand/collapse nodes, select, or drag to adjust structure \\
	AT37 & List & Infinite Scroll & Scroll to bottom to trigger loading more items \\
	AT38 & List & Item Click & Click list item to enter corresponding detail page \\
	AT39 & List & Drag Reorder & Drag list items to adjust display order \\
	\midrule
	
	\multicolumn{4}{l}{\textit{Feedback}} \\*
	AT40 & Drawer & Drawer Operations & Open/close side drawer for filtering or form operations \\
	AT41 & Modal & Modal Operations & Open/close modal dialog for form or content selection \\
	AT42 & Popconfirm & Confirmation Action & Trigger confirmation popover and execute confirm/cancel \\
	\midrule
	
	\multicolumn{4}{l}{\textit{Data Visualization}} \\*
	AT43 & Chart & Data Point Hover & Hover chart elements to display data tooltip \\
	AT44 & Chart & Legend Filtering & Click legend to toggle data series visibility \\
	AT45 & Chart & Zoom \& Pan & Box-select zoom or drag to pan for data details \\
	AT46 & Chart & Data Drill-down & Click chart elements to drill into finer-grained data \\
	AT47 & Map & Zoom \& Pan & Adjust map view via scroll zoom and drag pan \\
	AT48 & Map & Marker Interaction & Click map markers to view POI details \\
	AT49 & Map & Location Search & Search place names and navigate to location \\
	AT50 & Map & Route Planning & Set origin/destination and plan navigation route \\
	AT51 & Map & 3D View Adjustment & Adjust pitch and bearing angles via Ctrl+drag \\
	AT52 & Map & Street View & Enter street view, click to move, drag to look around \\
	AT53 & Panorama & Panoramic Browsing & Drag to rotate view, scroll to zoom details \\
	AT54 & Panorama & Hotspot Navigation & Click hotspot markers to switch panorama scenes \\
	AT55 & Model3D & Model Interaction & Drag to rotate, scroll to zoom, switch configurations \\
	\midrule
	
	\multicolumn{4}{l}{\textit{Media}} \\*
	AT56 & Video & Playback Control & Play/pause video, drag progress bar to seek \\
	AT57 & Video & Playback Settings & Adjust volume, speed, quality, subtitles, or fullscreen \\
	\midrule
	
	\multicolumn{4}{l}{\textit{Editor}} \\*
	AT58 & TextEditor & Text Editing & Edit rich text including formulas and special symbols \\
	AT59 & TextEditor & Format \& Content & Use toolbar for formatting or insert images/links/tables \\
	AT60 & Spreadsheet & Cell Editing & Select and edit spreadsheet cell contents \\
	AT61 & Spreadsheet & Formula Input & Enter formulas in cells for calculations \\
	AT62 & Spreadsheet & Drag Fill & Drag cell edges to batch fill data \\
	AT63 & Canvas & Element Manipulation & Select, move, and resize elements on canvas \\
	AT64 & Canvas & Node Connection & Drag from anchors to create connection lines \\
	
\end{longtable}
}

\subsection{Perception Types}
\label{appendix:perception_types}

Perception types ($\tau_{perc}$) characterize the visual understanding capabilities required to interpret UI states and content. These span four primary categories: \textit{Data Display}, \textit{Data Visualization}, \textit{Media}, and \textit{Editor}.

{
\small
\centering
\setlength{\tabcolsep}{5pt}
\begin{longtable}{@{}llp{3.1cm}p{5.9cm}@{}}
	\toprule
	\textbf{ID} & \textbf{Component} & \textbf{Perception Type} & \textbf{Description} \\
	\midrule
	\endfirsthead
	
	\toprule
	\textbf{ID} & \textbf{Component} & \textbf{Perception Type} & \textbf{Description} \\
	\midrule
	\endhead
	
	\midrule
	\endfoot
	
	\bottomrule
	\noalign{\vskip 6pt}
	\caption{Perception types for UI components.}
	\label{tab:perception_types} \\
	\endlastfoot
	
	\multicolumn{4}{l}{\textit{Data Display}} \\*
	PT01 & Tooltip & Content Understanding & Hover to obtain and understand tooltip text content \\
	PT02 & Tooltip & Type Recognition & Distinguish help, warning, error, and other tooltip types \\
	PT03 & Calendar & Date Marker Recognition & Identify dates with event or price markers on calendar \\
	PT04 & Calendar & Date State Recognition & Identify available, disabled, and selected date states \\
	PT05 & Card & Image Content Understanding & Recognize objects, scenes, and product types in card images \\
	PT06 & Card & Visual Feature Extraction & Extract color, material, and style features from card images \\
	PT07 & Card & Status \& Badge Recognition & Identify status markers (sold out, promo) and badge values \\
	PT08 & Carousel & Image Content Understanding & Recognize carousel image themes and promotional content \\
	PT09 & Carousel & Embedded Text Recognition & OCR text embedded within carousel images \\
	PT10 & Carousel & Position Awareness & Perceive current carousel position (n of total) \\
	PT11 & Image & Object Recognition & Identify specific objects and their categories in images \\
	PT12 & Image & Scene Understanding & Understand scene semantics (indoor/outdoor, purpose) \\
	PT13 & Image & Feature Extraction & Extract color, material, and style visual features \\
	PT14 & Image & Text Recognition & OCR text embedded within images \\
	PT15 & Image & Cross-image Comparison & Compare similarity across images, identify key differences \\
	PT16 & Collapse & Expansion State Awareness & Identify whether collapse panel is expanded or collapsed \\
	PT17 & Popover & Content Understanding & Understand comprehensive information in popover cards \\
	PT18 & List & Content Understanding & Understand list item content and extract key information \\
	PT19 & List & Sort State Awareness & Identify current list sorting method and order \\
	PT20 & Table & Data Understanding & Understand table data semantics and row-column relations \\
	PT21 & Table & Cell State Awareness & Identify row/cell states (highlighted, disabled, abnormal) \\
	PT22 & Table & Sort State Awareness & Identify current sort column and direction \\
	PT23 & Table & Cross-cell Understanding & Understand data relations across rows or columns \\
	PT24 & Tree & Hierarchy Understanding & Understand tree structure levels and parent-child relations \\
	PT25 & Tree & Node State Awareness & Identify node selection, expansion, and disabled states \\
	PT26 & Tree & Path Understanding & Understand complete path from root to target node \\
	\midrule
	
	\multicolumn{4}{l}{\textit{Data Visualization}} \\*
	PT27 & Chart & Value Reading & Accurately read specific values from visual positions \\
	PT28 & Chart & Trend Identification & Identify data trends (rising, falling, fluctuating, turning) \\
	PT29 & Chart & Multi-series Comparison & Compare differences and relations across data series \\
	PT30 & Chart & Proportion Understanding & Understand proportional relations in pie/donut charts \\
	PT31 & Chart & Extrema Identification & Find maximum/minimum points and corresponding positions \\
	PT32 & Chart & Anomaly Detection & Detect outliers, anomalies, and sudden changes in data \\
	PT33 & Chart & Legend Mapping & Understand correspondence between legend and data series \\
	PT34 & Map & Marker Position Understanding & Understand geographic positions and spatial relations \\
	PT35 & Map & Spatial Distribution Awareness & Identify distribution patterns, clusters, and coverage \\
	PT36 & Map & Route Understanding & Understand route path, distance, and estimated time \\
	PT37 & Map & Region Boundary Understanding & Understand regional divisions and administrative boundaries \\
	PT38 & Map & View State Awareness & Identify current view mode (2D, 3D, street) and tilt degree \\
	PT39 & Map & Street View Understanding & Understand buildings, roads, and environment in street view \\
	PT40 & Map & Street View Orientation & Understand directional orientation in street view \\
	PT41 & Map & 3D Occlusion Understanding & Understand occlusion relations between buildings in 3D \\
	PT42 & Panorama & Spatial Layout Understanding & Understand room layout and spatial structure in panorama \\
	PT43 & Panorama & Hotspot Recognition & Identify interactive hotspot marker positions \\
	PT44 & Panorama & Orientation Awareness & Understand current panorama viewing direction \\
	PT45 & Panorama & Environment Understanding & Understand decoration style, furniture in panorama \\
	PT46 & Model3D & Structure Understanding & Understand 3D model components and overall structure \\
	PT47 & Model3D & Material Perception & Identify surface materials and colors of model \\
	PT48 & Model3D & Proportion Perception & Understand relative proportions and size relations \\
	PT49 & Model3D & Viewpoint Recognition & Identify current viewing angle (front, side, back, etc.) \\
	\midrule
	
	\multicolumn{4}{l}{\textit{Media}} \\*
	PT50 & Video & Frame Content Understanding & Recognize content, objects, and scenes in video frames \\
	PT51 & Video & Playback State Awareness & Identify playback state (playing, paused, buffering) \\
	PT52 & Video & Progress Position Awareness & Understand current position within total duration \\
	\midrule
	
	\multicolumn{4}{l}{\textit{Editor}} \\*
	PT53 & Spreadsheet & Data Understanding & Understand spreadsheet data semantics and calculation logic \\
	PT54 & Spreadsheet & Format Semantics & Recognize conditional formatting and color-coded meanings \\
	PT55 & Canvas & Element Recognition & Identify shapes, lines, and text elements on canvas \\
	PT56 & Canvas & Relation \& Layout Understanding & Understand element connections, logic flow, and layout \\
	
\end{longtable}
}

\section{Prompts for Task Construction}
\label{appendix:prompts}

This appendix presents the prompts used in each phase of the \modelname\ construction pipeline. 

\subsection{Site Card Annotation Prompt}
\label{appendix:prompt_sitecard}

This prompt guides the annotation of site cards that document website capabilities. Given a target website URL, the annotation agent browses the site to identify functional modules, then enumerates complex execution items and perception items by analyzing the rendered HTML structure. The key insight is that complex UI components can be identified through their HTML signatures: for instance, a date range picker can be recognized by its characteristic DOM structure and attributes (e.g., \texttt{<input type="date">}, calendar container classes) without actually completing date selection.

\begin{tcolorbox}[colframe=gray!50!black, colback=gray!10!white, title=Site Card Annotation Prompt, breakable]
\small
\ttfamily

You are a frontend evaluation expert agent with browser automation capabilities. Your task is to visit the target website and systematically document its functions, UI components, and complex interaction points.

\medskip

INPUT:

- A target website URL to analyze

- A component taxonomy reference defining action types and perception types

\medskip

IDENTIFICATION METHOD:

\medskip

Identify complex UI components through HTML analysis:

- Inspect DOM structure, element tags, class names, and attributes

- Match against known component signatures from the taxonomy

- Record components based on their HTML presence, not execution success

\medskip

Examples of HTML-based identification:

- Date picker: \textless input type="date"\textgreater, calendar containers, date-range classes

- Map component: \textless div class="map-container"\textgreater, Leaflet/Google Maps scripts

- Video player: \textless video\textgreater tags, player control elements, playback-rate options

- Slider: \textless input type="range"\textgreater, slider track/handle elements

- Expandable section: aria-expanded attributes, collapse/accordion classes

\medskip

IMPORTANT: Your task is to IDENTIFY components through HTML inspection, not to COMPLETE complex operations. A component should be recorded if its DOM signature is present, regardless of whether you can fully operate it.

\medskip

VERIFICATION PRINCIPLES:

\medskip

1. Record What You Can Identify in HTML

~~- DOM structure matches known component pattern -> record

~~- Relevant classes/attributes present -> record

~~- No HTML evidence found -> do not record

\medskip

2. When In Doubt, Leave It Out

~~- Missing a real feature: causes incompleteness (acceptable)

~~- Recording a non-existent feature: causes downstream failure (unacceptable)

\medskip

STEP 1: IDENTIFY FUNCTIONS (F)

\medskip

Browse the website and identify user-facing functional modules.

\medskip

Naming: Use verb + domain-specific noun

- WRONG: "Search", "View Details"

- RIGHT: "Video Search", "Property Listing", "Course Reviews"

\medskip

Tags (combinable):

- [Core]/[Standard]: whether this is the primary draw of the site

- [Read]/[Write]: whether operations have persistent effects

\medskip

STEP 2: IDENTIFY EXECUTION ITEMS (A)

\medskip

For each function, identify complex UI components through HTML inspection.

\medskip

Two roles in tasks:

\medskip

1. Parameter Constraint - specifies how to perform an operation

~~- "Filter by price \$50-100, rating above 4.5, pet-friendly"

~~- "Play at 1.5x speed with subtitles enabled"

\medskip

2. Navigation Bridge - required step to access deeper capabilities

~~- "Enter detail page" -> enables further actions/perceptions

~~- "Expand fee breakdown" -> reveals hidden pricing

\medskip

Complexity Threshold:

- INCLUDE: Multi-criteria filtering, parameter configuration, map interactions, media control, complex forms

- EXCLUDE: Simple clicks, single inputs, basic navigation

\medskip

STEP 3: IDENTIFY PERCEPTION ITEMS (P)

\medskip

For each function, identify elements requiring visual understanding.

\medskip

Two roles in tasks:

\medskip

1. Information Output - agent must extract and return this

~~- "What is the monthly rent and deposit?"

~~- "Describe the price trend over 3 months"

\medskip

2. Filter Criteria - perception determines selection

~~- "Find pet-friendly apartments" (requires policy recognition)

~~- "Find items with recent price drops" (requires trend detection)

\medskip

Complexity Threshold:

- INCLUDE: Status recognition, structured data extraction, chart interpretation, spatial understanding

- EXCLUDE: Direct text reading, fixed-position extraction

\medskip

OUTPUT FORMAT:

\medskip

functions.csv:

~~id, name, tags, description

~~// description: core capabilities and typical use cases

\medskip

execution\_items.csv:

~~id, component, action\_type, associated\_functions, description

~~// description: location path and concrete usage scenario

\medskip

perception\_items.csv:

~~id, component, perception\_type, associated\_functions, description

~~// description: location path and what to extract or verify

\end{tcolorbox}

\subsection{Functional Affinity Assessment Prompt}
\label{appendix:prompt_affinity}

This prompt assesses pairwise cluster compatibility to produce the affinity matrix $\mathbf{M}$. Given two functional clusters as input, the model evaluates whether they can form coherent cross-site workflows and outputs an affinity score with justification.

\begin{tcolorbox}[colframe=gray!50!black, colback=gray!10!white, title=Functional Affinity Assessment Prompt, breakable]
\small
\ttfamily

You are assessing whether two functional clusters can form coherent cross-site workflows.

\medskip

INPUT: Two functional clusters C\_i and C\_j, each with member websites and their representative functions.

\medskip

TASK: Evaluate the plausibility of information flow scenarios:

- Can information from C\_i naturally serve as input for tasks in C\_j?

- Can information from C\_j naturally serve as input for tasks in C\_i?

- Do real user workflows commonly involve both clusters?

\medskip

RATING SCALE:

- HIGH: Frequent real-world workflows connect these clusters

~~Example: Academic Search + Code Hosting (researchers find papers then locate implementations)

- MEDIUM: Plausible but less common workflows exist

- LOW: No natural information flow

~~Example: Real Estate + Code Hosting (no coherent user scenario)

\medskip

OUTPUT: Affinity score (HIGH/MEDIUM/LOW) with brief justification.

\end{tcolorbox}

\subsection{Task Proposal Prompt}
\label{appendix:prompt_proposal}

This prompt generates task proposals that define the semantic structure of cross-site workflows. Given a sampled cluster subset $\mathcal{C}^*$ and the corresponding site cards, the model produces an array of task proposals, each specifying user intent, information flow pattern, and planned outputs. The proposals serve as semantic skeletons for downstream instantiation.

\begin{tcolorbox}[colframe=gray!50!black, colback=gray!10!white, title=Task Proposal Prompt, breakable]
\small
\ttfamily

You are designing cross-site tasks for an AI browser agent benchmark. Your goal is to create tasks that real users would genuinely want AI assistance with---NOT artificial capability tests.

\medskip

INPUT:

- Website cluster C containing functional clusters to use

- Site cards for each website, including: functions F, execution items A (complex UI operations like map dragging, date picker interaction), perception items P (visual understanding requirements like chart reading, table parsing)

\medskip

CORE PRINCIPLE: 

Imagine yourself as a user facing a tedious multi-site problem, thinking "I wish AI could handle this." That authentic feeling is the starting point for a good task. Do NOT design tasks just to test capabilities.

\medskip

INFORMATION FLOW PATTERNS:

Compose tasks using these primitives:

- Sequential (A -> B): A's output feeds B's input

- Parallel (A || B): Same goal across platforms for comparison

- Fan-out (A -> \{B1, B2\}): One source, multiple follow-ups

- Fan-in (\{A1, A2\} -> B): Multiple sources, unified processing

- Chain (A -> B -> C): Multi-hop information refinement

\medskip

AUTHENTICITY REQUIREMENTS:

- Task must address genuine user pain points (cross-platform switching, repetitive operations, scattered information)

- Manual completion should be tedious enough to warrant AI assistance

- Must require actual website operations (filtering, sorting, pagination, state verification), NOT just answerable by web search

\medskip

CROSS-SITE NECESSITY:

- Information flow between websites must be natural, not contrived

- Each website must provide unique, essential information

- Removing any website should break the task (Skip Test)

\medskip

DYNAMIC DEPENDENCY (CRITICAL):

All information passed between sites must be obtained during task execution, NOT preset in the task description.

\medskip

Reject these anti-patterns:

\medskip

1. Preset Keywords

~~WRONG: "Find 'Roman Architecture' courses on Coursera, then search 'Basilica' and 'Forum' on Getty Museum"

~~Problem: 'Basilica' and 'Forum' are preset, not discovered from Coursera

\medskip

2. Hypothetical Linking

~~WRONG: "Assuming CVPR is one of the conferences you found, search for hotels in San Francisco..."

~~Problem: "Assuming" indicates the input is preset by task designer, not dynamically obtained

\medskip

3. Isolated Outputs

~~WRONG: "Check flu activity level on CDC, find immunity-boosting ingredients on WebMD, then find recipes on FoodNetwork"

~~Problem: CDC output doesn't affect subsequent steps; removing it doesn't break the task

\medskip

GOOD EXAMPLES:

\medskip

1. "Find 5 Greek mythology artworks currently on display at Getty, note the artwork names and mythological figures, then find educational videos about EACH figure on Bilibili"

~~Why good: Must obtain specific artwork/figure names from Getty before knowing what to search on Bilibili

\medskip

2. "Check Saturday's precipitation probability on AccuWeather and AQI on EPA; if precipitation > 30\% OR AQI > 50, find indoor oven recipes, otherwise find outdoor BBQ recipes"

~~Why good: Weather data determines which recipe category to search---true conditional branching

\medskip

OUTPUT FORMAT:

Return a JSON array of task proposals:

[

~~\{

~~~~"task\_id": "PROP-001",

~~~~"user\_scenario": \{

~~~~~~"persona": "...",~ // Specific role, e.g., "ML researcher preparing a survey"

~~~~~~"situation": "...",~ // Concrete context triggering this task

~~~~~~"pain\_point": "...",~ // Why manual execution is tedious

~~~~~~"decision\_goal": "..."~ // Actionable outcome user seeks

~~~~\},

~~~~"information\_flow": \{

~~~~~~"pattern": "...",~ // One of: sequential, parallel,

~~~~~~~~~~~~~~~~~~~~~~~~ // fan-out, fan-in, chain, mixed

~~~~~~"description": "..."~ // How data transfers between sites

~~~~\},

~~~~"websites\_involved": ["site1.com", "site2.com", ...],

~~~~"function\_points": ["site1.com:F1", "site2.com:F3", ...],

~~~~"planned\_outputs": ["...", "..."]~ // What user expects to receive

~~\},

~~...~ // More proposals

]

\end{tcolorbox}

\subsection{Task Instantiation Prompt}
\label{appendix:prompt_instantiation}

This prompt transforms each task proposal into a concrete, executable benchmark instance. Given a single proposal and the relevant site cards, the model produces a detailed task specification including natural language description, structured output fields, and covered execution/perception items. The instantiation process binds abstract slots to concrete entities while ensuring complex interactions are triggered implicitly through goal-oriented descriptions.

\begin{tcolorbox}[colframe=gray!50!black, colback=gray!10!white, title=Task Instantiation Prompt, breakable]
\small
\ttfamily

You are instantiating a task proposal into a concrete, executable, and verifiable benchmark task.

\medskip

INPUT:

- A single task proposal with user scenario and information flow

- Site cards with functions F, execution items A and perception items P for involved websites

\medskip

GOAL-ORIENTED ELICITATION:

Express requirements through user goals, NOT UI operations. This ensures complex interactions are triggered implicitly while keeping task descriptions natural.

\medskip

WRONG: "Use the duration filter to select videos under 10 minutes"

RIGHT: "Find tutorial videos under 10 minutes"

\medskip

WRONG: "Scroll down to load more comments, then click expand"

RIGHT: "Check what users are discussing about this product"

\medskip

WRONG: "Click the expand button to show the full description"

RIGHT: "Verify if the video description mentions the original paper"

\medskip

WRONG: "Hover over the seller name to view reputation score"

RIGHT: "Find products from sellers with reputation above 4.8"

\medskip

OUTPUT FIELD DESIGN:

Every output field must serve one of three verification roles:

\medskip

1. User Value: Core information the user actually needs

~~Examples: product name, course title, paper abstract

\medskip

2. Condition Verification: Fields to validate filter criteria used in the task

~~CRITICAL: If task says "rating >= 4.5", output MUST include the rating

~~If task says "published within 2 years", output MUST include date

~~If task says "distance under 5 miles", output MUST include distance

\medskip

3. Provenance: URLs for source verification to prevent hallucination

~~Every claimed fact should be traceable to a source URL

\medskip

ROBUSTNESS GUIDELINES:

\medskip

1. Prefer Range Filtering over Superlatives

~~WRONG: "Find the most downloaded model" (sorting may not be supported)

~~RIGHT: "Find 5 models with downloads exceeding 10,000"

~~Reason: Superlatives require specific sorting features; range filters are more robust

\medskip

2. Prefer Relative Time over Absolute Dates

~~WRONG: "Find events on June 20, 2026" (task expires)

~~RIGHT: "Find events next weekend"

~~Reason: Absolute dates make tasks time-sensitive and eventually invalid

\medskip

3. Include Fallback Instructions for Dynamic Content

~~"If this weekend has no events, check next weekend"

~~"If the course is unavailable, find alternatives from the same instructor"

~~Reason: Web content changes; fallbacks ensure task remains completable

\medskip

4. Specify Quantities Explicitly

~~WRONG: "Find some highly-rated restaurants"

~~RIGHT: "Find 5 restaurants with rating above 4.5"

~~Reason: Explicit quantities enable objective evaluation

\medskip

5. Consolidate Output Requirements

~~Place all output requirements at the END of task description, not scattered throughout

~~WRONG: "Find books on Goodreads, note title and author. Then check reviews on StoryGraph, record the mood tags. Finally..."

~~RIGHT: "Find books on Goodreads... check reviews on StoryGraph... Output for each book: title, author, Goodreads rating, StoryGraph mood tags, and URLs for both platforms."

\medskip

TASK DESCRIPTION STYLE:

- First person ("I want...", "Help me...")

- Natural conversational tone, as if talking to an AI assistant

- 100-250 words

- Include context that implicitly triggers complex operations

\medskip

OUTPUT FORMAT:

\{

~~"task\_id": "TASK-001",

~~"source\_proposal": "PROP-001",~ // Reference to the source proposal

~~"task\_description": "...",~ // First-person natural language, 100-250 words

~~"information\_flow\_summary": "...",~ // e.g., "Site A -> Site B -> Site C"

~~"core\_outputs": [

~~~~\{

~~~~~~"id": "o1",~ // Unique identifier for this output field

~~~~~~"name": "...",~ // Field name, e.g., "listing\_price"

~~~~~~"type": "...",~ // One of: text, number, boolean, url, list

~~~~~~"source": "...",~ // Website this field comes from

~~~~~~"role": "..."~ // One of: user\_value, condition\_verify, provenance

~~~~\},

~~~~...

~~],

~~"covered\_points": [

~~~~\{

~~~~~~"point\_id": "...",~ // Format: website.com:Fx:Ay or website.com:Fx:Pz

~~~~~~"ability\_type": "...",~ // e.g., "date range filter", "scroll loading"

~~~~~~"trigger\_method": "...",~ // How task description triggers this ability

~~~~~~"output\_ref": "...",~ // Which output field (e.g., "o2") verifies execution

~~~~~~"verify\_method": "..."~ // How to verify, e.g., "check value >= threshold"

~~~~\},

~~~~...

~~],

~~"metadata": \{

~~~~"websites\_involved": ["site1.com", "site2.com", ...],

~~~~"action\_points": ["site1.com:F1:A2", ...],~ // Covered execution items

~~~~"perception\_points": ["site2.com:F3:P1", ...]~ // Covered perception items

~~\}

\}

\end{tcolorbox}
\section{A Complete Task Instance}
\label{appendix:task_example}

This appendix presents a real task instance from \modelname\ together with the websites it spans, its covered points, and the automatically generated rubric. This instance covers 17 execution and 8 perception points, above the benchmark average of 7 and 4, and was chosen to illustrate all three dimensions (cross-site workflow, complex actions, and perception) at once.

\paragraph{Why shortcuts cannot pass.} Task descriptions are phrased as user goals rather than UI instructions, so no method is prescribed; instead, the rubric enforces that the required operations are actually reflected in the answer. Each required execution and perception point maps to a critical leaf whose claims must be supported by cited source pages, so an answer that merely looks right, whether guessed, retrieved through a plain search query, or reached via a direct URL, fails verification. Because tasks chain sites such that each step's output feeds the next step's input, skipping an early operation starves every downstream step. During task validation (\S\ref{subsec:task_inst}), annotators also executed each task on a live browser and confirmed that the target operations admit no reliable shortcut.

\begin{tcolorbox}[colframe=gray!50!black, colback=gray!10!white, title=Task Description, breakable]
\small
I'm writing a sci-fi script about AI ethics and want to systematically gather creative source material. First, go to IMDb and search for sci-fi films with theme tags including ``artificial intelligence'' or ``dystopia,'' with ratings above 7.5. Select 5 classic works and record the title, director, IMDb rating, and detail-page link for each.

\medskip
Then, search these 5 films on Letterboxd. For each film, check the community rating and top user reviews (prioritize highly liked reviews; if the page cannot be reliably sorted by likes, select the 3 most relevant visible reviews and state your selection criteria), and extract keywords describing commonly discussed narrative techniques (e.g., ``multi-thread narrative,'' ``time loop,'' ``first-person perspective'').

\medskip
Next, on Goodreads, try to find the original novels corresponding to these films or books from the same IP/franchise. Prioritize books with ratings above 4.0 and more than 1,000 reviews, and aim to find 3 books. If fewer than 3 are available, keep the results found and clearly state the gap. Record the title, author, Goodreads rating, philosophical topics mentioned in popular reviews (e.g., ``free will,'' ``the nature of consciousness''), and link.

\medskip
Then, go to Semantic Scholar and search for ``AI ethics'' or ``machine consciousness.'' Sort by citation count and find 5 papers with more than 100 citations (if fewer than 5 exist, record the actual number found and explain). Record the title, authors, citation count, publication year, abstract, and paper link.

\medskip
Finally, on YouTube, find in-depth analysis videos for any 2 of the 5 films (e.g., ``[Film] review,'' ``[Film] analysis''), with each video longer than 15 minutes and having more than 100,000 views; find 1 video per film. If a specific film has insufficient results, you may switch to another film among the 5 and expand search keywords (e.g., ``ending explained,'' ``deep dive''). Record the video title, channel name, view count, duration, and link.

\medskip
Output format:
\begin{itemize}[leftmargin=1.5em, label=-, nosep]
  \item IMDb films (title, director, rating, link)
  \item Letterboxd info (community rating, narrative-technique keywords)
  \item Goodreads books (title, author, rating, philosophical topics, link)
  \item Papers (title, authors, citation count, year, abstract, link)
  \item YouTube videos (title, channel, views, duration, link)
\end{itemize}

\medskip
If any step encounters a login wall or regional restriction, record the limitation and continue with the remaining steps.
\end{tcolorbox}

\paragraph{Websites involved.} IMDb $\rightarrow$ Letterboxd $\rightarrow$ Goodreads $\rightarrow$ Semantic Scholar $\rightarrow$ YouTube.

\needspace{7\baselineskip}
\paragraph{Covered execution points (17 total; selected examples).}
\begin{itemize}[leftmargin=1.5em, label=-, nosep]
  \item \texttt{imdb.com:F9:A19} -- Rating range slider (filter $\geq$ 7.5)
  \item \texttt{imdb.com:F9:A11} -- Multi-item checkbox selection (Sci-Fi genre filter)
  \item \texttt{letterboxd.com:F3:A8} -- Tab panel switch (navigate to Reviews tab)
  \item \texttt{semanticscholar.org:F1:A2} -- Column sorting (sort by citation count)
  \item \texttt{youtube.com:F1:A22} -- Search-filtered selection (duration/view count filter)
\end{itemize}

\paragraph{Covered perception points (8 total; selected examples).}
\begin{itemize}[leftmargin=1.5em, label=-, nosep]
  \item \texttt{letterboxd.com:F3:P6} -- Content understanding (extract narrative technique keywords from reviews)
  \item \texttt{goodreads.com:F2:P5} -- Status \& badge recognition (identify rating and review count)
  \item \texttt{youtube.com:F1:P4} -- Video playback state awareness (identify duration and view count from cards)
\end{itemize}

\paragraph{Generated rubric (abbreviated).}
\begin{tcolorbox}[colframe=gray!50!black, colback=gray!10!white, breakable]
\small
\ttfamily
\begin{verbatim}
Root: Overall Task Completion
|-- Step 1 [non-critical]: IMDb Filtering
|   |-- [action] Rating slider >= 7.5 applied (critical)
|   |-- [action] Sci-Fi genre filter applied (critical)
|   `-- [perception] 5 film titles correctly extracted (critical)
|-- Step 2 [non-critical]: Letterboxd Review Extraction
|   |-- [action] Reviews tab navigated (critical)
|   `-- [perception] Narrative technique keywords extracted (non-critical)
|-- Step 3 [non-critical]: Goodreads Book Search
|   |-- Rating >= 4.0 and reviews > 1,000 verified (critical)
|   `-- [perception] Philosophical topic keywords extracted (non-critical)
|-- Step 4 [non-critical]: Semantic Scholar Paper Search
|   |-- [action] Sort by citation count applied (critical)
|   `-- Citation count >= 100 verified for each paper (critical)
`-- Step 5 [non-critical]: YouTube Video Search
    |-- View count > 100k verified (critical)
    `-- Duration >= 15 min verified (critical)
\end{verbatim}
\end{tcolorbox}

\section{Quality Control Criteria}
\label{appendix:quality_control}

Each instantiated task undergoes manual quality control by twelve annotators. Tasks are first grouped by function sets and websites, with redundant ones removed within each group. Each task is then reviewed against five criteria: rationality, complexity, executability, evaluability, and topic safety. Annotators also execute each task on the live websites, followed by agent-based testing.

The finalized benchmark contains 420 task instances with an average of $|\mathcal{E}_{act}| = 7$ execution items and $|\mathcal{E}_{perc}| = 4$ perception items per task, involving up to 5 clusters and 14 websites.

\paragraph{Annotation protocol.}
All annotators are CS graduate students, paid \$1 per task instance; each completed a training session and at least five pilot annotations before working on the full dataset. Annotation proceeds in two stages with separate teams: eight annotators carried out proposal screening (\S\ref{subsec:task_proposal}), judging each generated proposal on whether the user intent is meaningful and whether the cross-site information is dynamically obtained, and twelve annotators carried out the task-instantiation quality control described above, reviewing every instantiated task against the five criteria and personally executing it on a live browser to confirm feasibility.

\paragraph{Site card error analysis.}
To quantify the reliability of the automated site-card construction (\S\ref{subsec:sitecard}), we randomly sampled 36 of the 108 site cards, spanning 18 domains (academic search, code hosting, e-commerce, healthcare, maps, real estate, travel, etc.), and manually checked every entry against the live site, conservatively counting anything that could not be verified as an error. Results are shown in Table~\ref{tab:sitecard_error}.

\begin{table}[h]
  \centering
  \renewcommand{\arraystretch}{1.1}
  \begin{tabular}{lcc}
    \toprule
    \textbf{Item type} & \textbf{Correct / Total} & \textbf{Accuracy} \\
    \midrule
    Function   & 371 / 382   & 97.1\% \\
    Perception & 602 / 640   & 94.1\% \\
    Action     & 756 / 852   & 88.7\% \\
    \midrule
    Overall    & 1,729 / 1,874 & 92.3\% \\
    \bottomrule
  \end{tabular}
  \caption{Manual error analysis of automatically constructed site cards on a random sample of 36 of the 108 cards across 18 domains.}
  \label{tab:sitecard_error}
\end{table}

The cards are accurate overall (92.3\%): site functions and perception items are near-ceiling (97.1\% / 94.1\%), and errors concentrate in action items (88.7\%). These errors are not invented site functionality but over-predictions of plausible fine-grained UI controls (e.g., dropdowns, collapsible panels, map widgets) that the automatic builder anticipated but that only become visible after interactions it did not perform. A wrong card entry also cannot propagate into the benchmark: every task is manually executed on the live site before it is admitted, so any task relying on an entry that does not actually exist fails this check and is dropped.

\section{Information Flow Patterns}
\label{appendix:info_pattern}

We represent the information flow $\phi_{prop}$ as a directed acyclic graph (DAG) over workflow steps. Formally, $\phi_{prop} = (V, E)$ where each step $v_i = (w_i, f_i, o_i) \in V$ invokes function $f_i$ on website $w_i$ to produce output $o_i$, and each edge $(v_i, v_j) \in E$ indicates that $o_i$ serves as input to $v_j$. To promote diversity, we provide the language model with five primitive patterns (listed below) and let it compose tasks by freely combining them. For example, a workflow can merge a sequential step with a parallel branch and a fan-in aggregation, represented symbolically as
\[
v_1 \to (v_2 \parallel v_3) \to v_4,
\]
where $v_1$ triggers $v_2 \parallel v_3$, both feeding into $v_4$. By exploring such combinations, we generate workflows $\phi_{prop}$ that are plausible across websites and structurally varied.

\begin{table}[ht]
	\centering
	\small
	\setlength{\tabcolsep}{3.8pt}
	\begin{tabular}{llp{3.6cm}}
		\toprule
		\textbf{Pattern} & \textbf{Structure} & \textbf{Example} \\
		\midrule
		Sequential & $v_i \to v_j$ & Find paper on arXiv, then locate code on GitHub \\
		Parallel & $v_i \parallel v_j$ & Cross-platform price comparison \\
		Fan-out & $v_i \to \{v_j\}$ & Obtain recommendations, verify each on separate sites \\
		Fan-in & $\{v_i\} \to v_j$ & Aggregate reviews from multiple sources for decision \\
		Chain & $v_i \to v_j \to v_k$ & News article $\to$ company website $\to$ financial tracker \\
		\bottomrule
	\end{tabular}
	\caption{Primitive cross-site information flow patterns.}
	\label{tab:patterns}
\end{table}
\section{Prompt for Evaluation Script Generation}
\label{appendix:eval_detail}

This appendix presents the prompt used for Evaluation Script Generation, corresponding to the methodology described in \S\ref{sec:4_2}. This prompt instructs the LLM to translate the structured task metadata into executable Python verification scripts.

\begin{tcolorbox}[ colframe=gray!50!black, colback=gray!10!white, 
    title=Evaluation Script Generation Prompt, 
    breakable
]
\small
\ttfamily

You are a skilled evaluation annotator. Please write a Python evaluation script based on the Mind2Web 2 framework for a new task, using the following task information and reference code.

\medskip

\textbf{\#\#\# New Task Information}

\{json.dumps(context, indent=2, ensure\_ascii=False)\}

\medskip

\textbf{\#\#\# Core Requirements}

1. \textbf{Focus on covered\_points}: Design evaluation points based on the \texttt{covered\_points} field.
\begin{itemize}[leftmargin=1.5em, label=-, nosep]
    \item \texttt{point\_id} examples: "ikea.com:F2:A8" (Action) or "ikea.com:F9:P6" (Perception).
    \item \textbf{Action Node} \texttt{desc} format: \texttt{"<complex action: [point\_id]> ..."}
    \item \textbf{Perception Node} \texttt{desc} format: \texttt{"<complex perception: [point\_id]> ..."}
\end{itemize}

2. \textbf{Strict Critical Node Logic (Error Prevention)}:
\begin{itemize}[leftmargin=1.5em, label=-, nosep]
    \item \textbf{Cause of Error}: The framework stipulates that if a parent node is \texttt{critical=True}, all its child nodes must also be \texttt{critical=True}.
    \item \textbf{Design Recommendation}: To avoid errors and allow for partial scoring, \textbf{please set top-level container nodes (e.g., step1, step2) to \texttt{critical=False}}. Only set leaf nodes (specific verification nodes) to \texttt{critical=True} if the failure of that step implies total task failure.
\end{itemize}

3. \textbf{Do not deviate from the task}: Do not add irrelevant evaluation points not mentioned in the task description.

\medskip

\textbf{\#\#\# Code Writing Specifications (API Whitelist)}

\textbf{You must strictly adhere to the following APIs; do not invent non-existent methods:}

1. \textbf{Allowed Methods}:
\begin{itemize}[leftmargin=1.5em, label=-, nosep]
    \item \texttt{evaluator.initialize(...)}: Initialize, returns the root node.
    \item \texttt{evaluator.add\_sequential(...)}: \textbf{Can} be used (Note: not \texttt{add\_sequence}). Used to add sequential execution container nodes.
    \item \texttt{evaluator.add\_parallel(...)}: Used to add parallel execution container nodes.
    \item \texttt{evaluator.add\_leaf(...)}: Add a leaf node.
    \item \texttt{evaluator.add\_custom\_node(...)}: Pass in a boolean result node directly.
    \item \texttt{evaluator.extract(...)}: Extract information.
    \item \texttt{evaluator.verify(...)}: Verification logic.
    \item \texttt{evaluator.get\_summary()}: Return the result.
\end{itemize}

2. \textbf{Advanced Technique (verify with node=None)}:
\begin{itemize}[leftmargin=1.5em, label=-, nosep]
    \item If you need to obtain a verification result (True/False) in the Python code to determine subsequent logic, but do not want to draw this node on the evaluation tree, use \texttt{await evaluator.verify(node=None, ...)}.
\end{itemize}

3. \textbf{Strictly Prohibit Hallucinations (Negative Constraints)}:
\begin{itemize}[leftmargin=1.5em, label=-, nosep]
    \item \textbf{Prohibited Usage}: \texttt{add\_sequence} (missing 'ial'), \texttt{add\_serial}, \texttt{update\_node\_result}.
    \item \textbf{Prohibited Attribute Access}: The \texttt{Evaluator} object does not have \texttt{.logger} or \texttt{.answer} attributes.
    \item \textbf{Correct Practice}: \texttt{logger} and \texttt{answer} are arguments of the \texttt{evaluate\_answer} function; use the argument variables directly (e.g., \texttt{logger.info(...)}).
\end{itemize}

4. \textbf{Data Structures}:
\begin{itemize}[leftmargin=1.5em, label=-, nosep]
    \item Use \texttt{pydantic.BaseModel} to define extraction structures.
    \item The \texttt{evaluate\_answer} function signature must remain complete (containing all arguments: \texttt{client}, \texttt{answer}, \texttt{cache}, \texttt{semaphore}, \texttt{logger}, \texttt{model} etc.).
\end{itemize}

\medskip

\textbf{\#\#\# Reference Code Template}

Please imitate the structure, import style, and verification logic of the following code:

\texttt{-{}-{}-}

\{reference\_code\}

\texttt{-{}-{}-}

Please output the Python code directly. Do not include Markdown code block markers, and ensure the code is directly executable.

\end{tcolorbox}
\section{Validation of the Verifiable Agent-as-a-Judge Framework}
\label{appendix:judge_validation}

Because our results depend on the verifiable agent-as-a-judge framework, we validate it in three ways: agreement with human graders, validity of the generated rubrics, and robustness to the choice of judge backbone. All analyses use 50 tasks uniformly sampled across task clusters and across the number of sites involved.

\paragraph{Agreement with human graders.}
On each of the 50 tasks we scored two answers, the human baseline's and the best agent's (Comet), and had a human grader and the judge independently mark each answer pass/fail, giving $50 \times 2 = 100$ verdict pairs. The two agree on 96 of 100 verdicts (96\% agreement; judge--human Cohen's $\kappa = 0.84$, ``almost perfect'' on the Landis--Koch scale). The confusion matrix is shown in Table~\ref{tab:judge_agreement}. All four disagreements run the same way: the judge accepted an answer the human grader rejected, never the reverse, so the judge is at worst slightly lenient and is not biased against agents. Where the judge failed an answer, human graders agreed that the answer omitted a required filter, skipped a downstream site, or provided no source evidence.

\begin{table}[h]
  \centering
  \renewcommand{\arraystretch}{1.1}
  \begin{tabular}{lcc}
    \toprule
     & \textbf{Judge: pass} & \textbf{Judge: fail} \\
    \midrule
    \textbf{Human: pass} & 13 & 0 \\
    \textbf{Human: fail} & 4  & 83 \\
    \bottomrule
  \end{tabular}
  \caption{Judge--human agreement on 100 pass/fail verdicts (50 tasks $\times$ 2 answers). Agreement is 96\% (Cohen's $\kappa = 0.84$).}
  \label{tab:judge_agreement}
\end{table}

\paragraph{Rubric validity.}
Separately, annotators rated whether each task's rubric tree (the checklist the judge scores against) faithfully captures the task: 48 of 50 (96\%) were rated correct. The remaining two revealed script bugs that we fix in the released evaluator: one was too lenient on missing/\texttt{None} fields, and one under-covered a repeated requirement.

\paragraph{Robustness across judge backbones.}
To check that our conclusions do not depend on the specific judge model, we re-scored the same answers with five judge backbones: GPT-4o (used in the paper), Claude-Opus-4.8, GLM, Qwen3-235B, and DeepSeek-V4-Flash. Table~\ref{tab:judge_robustness} reports Partial Completion for three evaluated systems (Comet, the Human baseline, and the Browser-Use~+~DeepSeek-V4-Flash agent); we use Partial Completion because full-success rates are single-digit and too sparse to compare reliably across judges. Absolute scores shift by a few points across judges, but the ranking Comet $>$ Human $>$ DeepSeek holds under every judge, so the paper's conclusions are unchanged.

\begin{table}[h]
  \centering
  \renewcommand{\arraystretch}{1.1}
  \begin{tabular}{lccc}
    \toprule
    \textbf{Judge backbone} & \textbf{Comet} & \textbf{Human} & \textbf{DeepSeek} \\
    \midrule
    \rowcolor{gray!12}
    GPT-4o (paper)     & 48.0 & 35.0 & 25.0 \\
    Claude-Opus-4.8    & 46.5 & 27.8 & 21.2 \\
    GLM                & 39.3 & 25.9 & 22.2 \\
    Qwen3-235B         & 46.5 & 32.2 & 26.7 \\
    DeepSeek-V4-Flash  & 44.4 & 30.8 & 22.5 \\
    \bottomrule
  \end{tabular}
  \caption{Partial Completion (\%) of three evaluated systems under five different judge backbones. The system ranking is preserved under every judge.}
  \label{tab:judge_robustness}
\end{table}

\end{document}